\documentclass[lettersize,journal]{IEEEtran}
\usepackage{amsmath,amsfonts}
\usepackage{algorithmic}
\usepackage{algorithm}
\usepackage{array}
\usepackage[caption=false,font=normalsize,labelfont=sf,textfont=sf]{subfig}
\usepackage{textcomp}
\usepackage{stfloats}
\usepackage{url}
\usepackage{verbatim}
\usepackage{graphicx}
\usepackage{cite}
\usepackage{subfig}
\usepackage{xcolor}
\usepackage[citecolor=blue, colorlinks]{hyperref}
\usepackage{multirow}

\newcommand{\revise}[1]{\textcolor{black}{#1}}

\begin{document}

\title{Boosting 3D Object Detection with Semantic-Aware Multi-Branch Framework}

\author{Hao Jing, Anhong Wang, Lijun Zhao, Yakun Yang,  Donghan Bu, Jing Zhang, Yifan Zhang, and Junhui Hou

\thanks{This work was supported by the National Natural Science Foundation of China (No. 62072325, No. U23A20314, No. 62202323), Industrial Application of Shanxi Provincial Technology Innovation Center (IVASXTIC2022), Shanxi Province 'Reveal the List' Major Project (202301156401007), Shanxi Scholarship Council of China (2024-130), Taiyuan Key Core Technology Research 'Reveal the List' Project (2024TYJB0128, 20240027), the NSFC Excellent Young Scientists (No. 6242211), Hong Kong Innovation and Technology (ITS/164/23, MHP/117/21). \textit{(Corresponding authors: Anhong Wang; Lijun Zhao; Yifan Zhang.)}}
\thanks{Hao Jing, Anhong Wang, Lijun Zhao, Yakun Yang, Donghan Bu, and Jing Zhang are with the School of Electronic Information Engineering, Taiyuan University of Science and Technology, No. 66 Waliu Road, Taiyuan 030024, China. (e-mail: b20201591001@stu.tyust.edu.cn; ahwang@tyust.edu.cn; 2019010@tyust.edu.cn; yyk@tyust.edu.cn; b202115310021@stu.tyust.edu.cn; 2023025@tyust.edu.cn).}
\thanks{Yifan Zhang and Junhui Hou are with the Department of Computer Science, City University of Hong Kong, Kowloon Tong, Hong Kong, China (e-mail: yzhang3362-c@my.cityu.edu.hk; jh.hou@cityu.edu.hk).}
}

\markboth{Journal of \LaTeX\ Class Files,~Vol.~14, No.~8, August~2021}%
{Shell \MakeLowercase{\textit{et al.}}: A Sample Article Using IEEEtran.cls for IEEE Journals}


\maketitle

\begin{abstract}
In autonomous driving, LiDAR sensors are vital for acquiring 3D point clouds, providing reliable geometric information. However, traditional sampling methods of preprocessing often ignore semantic features, leading to detail loss and ground point interference in 3D object detection. To address this, we propose a multi-branch two-stage 3D object detection framework using a Semantic-aware Multi-branch Sampling (SMS) module and multi-view consistency constraints. The SMS module includes random sampling, Density Equalization Sampling (DES) for enhancing distant objects, and Ground Abandonment Sampling (GAS) to focus on non-ground points. The sampled multi-view points are processed through a Consistent KeyPoint Selection (CKPS) module to generate consistent keypoint masks for efficient proposal sampling. The first-stage detector uses multi-branch parallel learning with multi-view consistency loss for feature aggregation, while the second-stage detector fuses multi-view data through a Multi-View Fusion Pooling (MVFP) module to precisely predict 3D objects. The experimental results on the KITTI dataset and Waymo Open Dataset show that our method achieves excellent detection performance improvement for a variety of backbones, especially for low-performance backbones with simple network structures. The code will be publicly available at \url{https://github.com/HaoJing-SX/SMS.}
\end{abstract}


\section{Introduction}
\IEEEPARstart{L}{idar} sensors are pivotal devices for acquiring 3D point clouds in traffic scenarios. They reliably provide sparse geometric information due to the distinctive geometric features of 3D point clouds and their immunity to weather and lighting conditions. In autonomous driving applications, 3D object detection from point clouds plays a crucial role in scene understanding. Given that point clouds are characterized by uncertain size and non-uniform density, most 3D detectors employ specific sampling methods during data preprocessing to transform the raw point cloud into an accessible and structured representation\cite{yang20203dssd}. Generally, these preprocessed point clouds can better support 3D object detectors than the original data, enabling accurate predictions of each object's category, position, and 3D bounding box.

In 3D object detection, sampling methods for data preprocessing can be classified into two categories: grid-based methods\cite{deng2021voxel,yan2018second,zhou2018voxelnet,zhang2024stemd} and point-based methods \cite{yang20203dssd,shi2019pointrcnn,zhang2022not,chen2022sasa}. Grid-based sampling methods often convert point clouds into 3D voxels through voxelization sampling, which can lead to the loss of fine-grained information and thus limit the performance of 3D object detection. In contrast, point-based sampling methods apply random sampling directly on raw points to generate a fixed number of input point clouds. However, when downsampling is required to reduce the point count, point-based sampling methods that do not consider intrinsic semantics of point clouds may lose some crucial object points in low-density remote areas, thereby affecting detection accuracy. Furthermore, both approaches retain a significant number of ground points, resulting in redundant computations and potentially severe interference during detection. To address these issues, this paper proposes leveraging semantic information of point clouds to improve preprocessing sampling for 3D object detection, enabling the input point clouds to better focus on detection objects.

\begin{figure*}[!htb]
    \centering
    \includegraphics[width=5.7in]{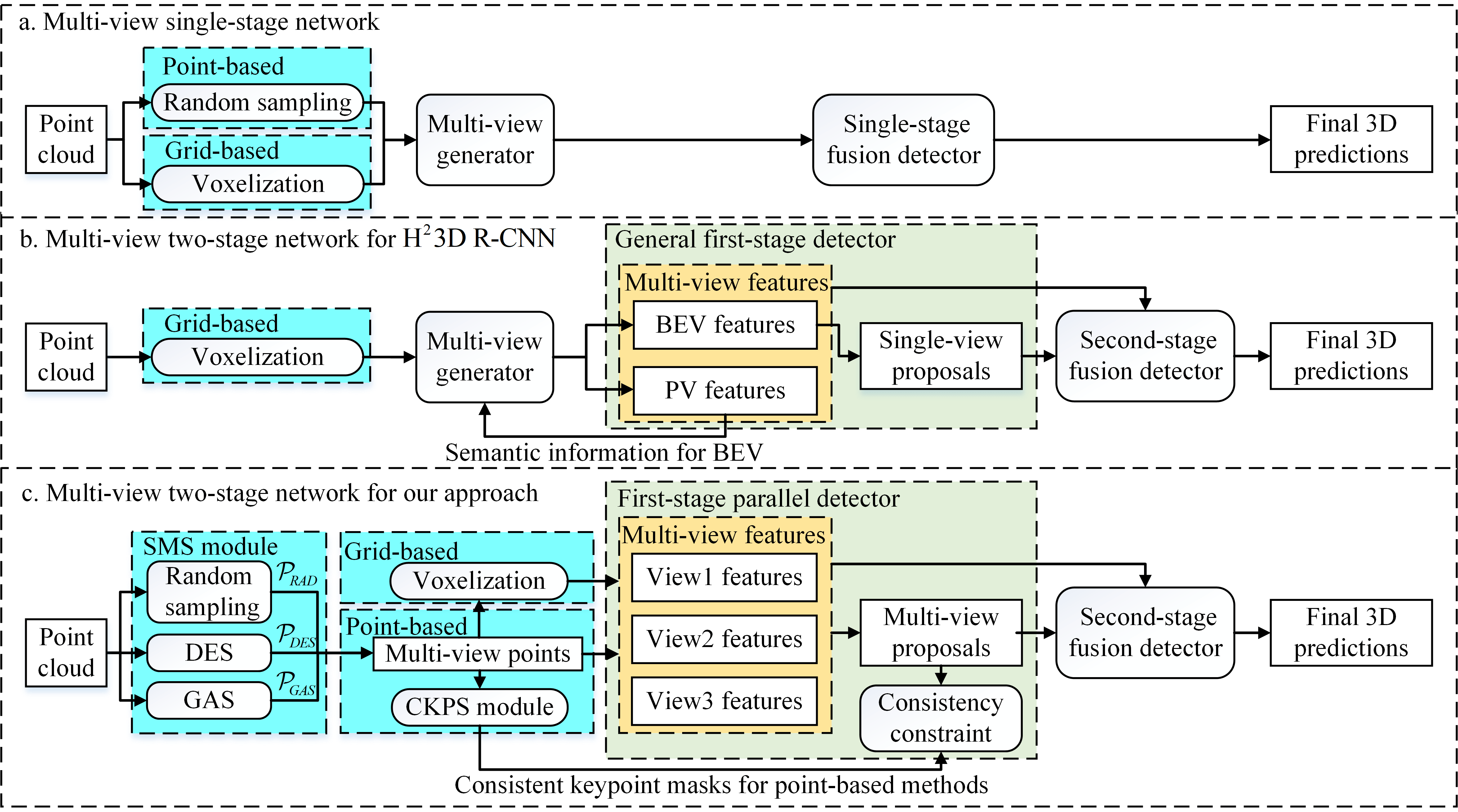} 
    \caption{The comparison between our approach and existing multi-view frameworks of 3D object detection.}
    \label{fig:figure1}
    \vskip -10pt
\end{figure*}

Existing 3D object detection frameworks can be divided into two classes: single-stage networks \cite{yang20203dssd,zhang2022not,zheng2021cia,ma2022cg,zheng2021se,li2023tinypillarnet} and two-stage networks \cite{deng2021voxel,shi2019pointrcnn,shi2020pv,fan2023hcpvf,qian2022badet,mao2021pyramid}. Generally, two-stage networks exhibit higher detection accuracy. In a typical two-stage network pipeline, the first-stage detector extracts semantic features from input point clouds to generate proposals and then the second-stage detector refines local semantic features of these proposals through a Region-of-Interest (RoI) pooling to improve detection accuracy \cite{shi2019pointrcnn}. In comparison, single-stage networks eliminate the second-stage detector and focus on optimizing the feature extraction network and detection head within the first stage, which typically results in limited performance. As depicted in Fig. \hyperref[fig:figure1]{\ref{fig:figure1}(a)}, multi-view single-stage networks \cite{fazlali2022versatile,zhou2020end,yang2020multi} employ multi-view generators to extract semantic features from different perspectives of point clouds, thus achieving significant performance improvements over traditional single-stage networks. However, these multi-view single-stage networks overlook the unique semantics of each view and fuse multi-view features prematurely, causing lower performance than traditional two-stage networks. To address this issue, the multi-view two-stage network H$^2$3D R-CNN \cite{deng2021multi} (see Fig. \hyperref[fig:figure1]{\ref{fig:figure1}(b)}) learned unique semantic features from multiple views in parallel during the first stage and then fused these multi-view features through a cross-view gating mechanism in the second stage. However, this approach lacks both separate supervision of each view’s proposals and unified supervision to ensure multi-view proposal consistency, which limits its effectiveness in feature aggregation across multiple views.

Inspired by the above analysis, this paper proposes a multi-branch two-stage 3D object detection framework based on Semantic-aware Multi-branch Sampling (SMS) and multi-view consistency constraints, as shown in Fig. \hyperref[fig:figure1]{\ref{fig:figure1}(c)}. In the data preprocessing stage, in addition to random sampling, we introduce two additional semantic-aware sampling branches to generate multi-view point clouds with distinct semantic emphases, including Density Equalization Sampling (DES) and Ground Abandonment Sampling (GAS). DES formulates density sampling rules according to \revise{the semantic information of regional density} to appropriately increase the point count for detection objects in low-density remote areas. GAS filters non-ground object points using \revise{the semantic information of local height} so as to focus on to-be-detected key objects. After these multi-branch sampling operations, the first-stage parallel detector extracts semantic features from each branch of multiple views to predict proposals independently. To achieve better multi-view feature aggregation, we propose to impose proposal-level consistency constraints based on the foreground sampling of multi-view proposals. Specifically, to address the challenges of misalignment, high density and overlap for point-based multi-view proposals, we design a Consistent KeyPoint Selection (CKPS) module to generate the spatially aligned and uniform consistent keypoint masks. This module can assist foreground sampling in efficiently obtaining multi-view proposals corresponding to consistent foreground points. In the second-stage fusion detector, we introduce the Multi-View Fusion Pooling (MVFP) module to blend points, features, and proposals simultaneously, so that the obtained hybrid features provide comprehensive information for final 3D predictions.

Our contributions can be summarized in the following four aspects: (1) We design a pluggable SMS module with three branches, including random sampling, DES, and GAS, which can be applied to the data preprocessing stage of various two-stage 3D object detectors without changing the backbone structures. (2) According to the backbone's data type, raw point clouds can be transformed into multi-view points or voxels with distinct semantic emphases through SMS-based data preprocessing, effectively preparing the input data for subsequent object detection. (3) Building on the sampled multi-view points, we introduce multi-view consistency constraints within a universal multi-branch two-stage 3D object detection framework, leveraging first-stage multi-branch parallel training and second-stage fusion learning to achieve effective multi-view feature aggregation. (4) We evaluate the proposed framework on the KITTI dataset and Waymo Open Dataset, demonstrating state-of-the-art 3D object detection performance.

\section{Related Works}
\subsection{\revise{Data Preprocessing and Sampling in 3D Object Detection}}
To improve the performance of 3D object detection, recent research focused on point cloud data preprocessing usually involves two steps: data augmentation and data processing. For data augmentation, Ground Truth (GT) sampling was thoroughly explored on spatial non-overlap \cite{yan2018second}, ground alignment \cite{zhu2019class}, and contextual surrounding \cite{lee2023resolving}. A series of methods also employed distance-aware \cite{wu2023transformation}, shape-aware \cite{zheng2021se} and occlusion-driven \cite{zhang2023glenet} augmentations to address challenges in sparse and incomplete instances. For data processing, the significant sampling module can be broadly divided into two categories: grid-based methods and point-based methods. \revise{In grid-based sampling methods \cite{zhou2018voxelnet,yan2018second,deng2021voxel,xiao2023balanced,yu2023pipc}, voxelization sampling transformed raw point clouds into 3D voxels with the deficient fine-grained information. In contrast, point-based sampling methods \cite{shi2019pointrcnn,shi2020point,yang20203dssd,chen2022sasa,zhang2023unleash} reduced the raw point count through random sampling to obtain a fixed number of input points, resulting in information loss for distant objects with fewer points. Both grid-based and point-based methods are subject to interference from the remaining ground points. The dynamic voxelization proposed by MVF \cite{zhou2020end} only mitigated the fine-grained information deficiency but failed to overcome ground interference. To address this, our SMS-based data preprocessing combines distant object enhancement and ground abandonment within a multi-branch sampling module, which is adaptable to both grid-based and point-based backbones.}

\subsection{\revise{Semantics in 3D Object Detection}}
\revise{The semantics of 3D object detection include essential attributes like position, size, orientation, category, and other characteristics, which aid in understanding and analyzing real-world 3D scenes. In multi-modal methods, EPNet \cite{huang2020epnet} and EPNet++ \cite{liu2022epnet++} leveraged unique 2D image information (e.g., color and texture) to enrich object semantics, thus enhancing feature extraction and 3D prediction. In comparison, point cloud-based methods primarily focus on learning geometric information to extract semantic features, such as point-wise features \cite{shi2019pointrcnn,yang20203dssd} and grid-wise features \cite{zhou2018voxelnet,yan2018second,deng2021voxel}. For various feature formats derived from geometric semantics, VoteNet \cite{qi2019deep} introduced voting features, and H3DNet \cite{zhang2020h3dnet} designed geometric primitives, both of which achieved significant performance improvements. To provide additional semantic supervision, HyperDet3D \cite{zheng2023learning} learned point-wise features for specific and agnostic scenes to obtain parameterized prior knowledge, while SCGNet \cite{dong2023semantic} established a semantic-context graph based on proposal categories and contextual relationships. Building upon the abstract semantics from geometric information, PDV \cite{hu2022point} integrated regional density features in RoI pooling to refine proposals, and ProposalContrast \cite{yin2022proposalcontrast} mitigated ground interference by abandoning ground plane points determined by local height. Inspired by these methods, we propose the SMS module that synthetically utilizes the abstract semantic information of regional density and local height to generate multi-view input points with distinct semantic emphases. Moreover, this sampling strategy is not confined to our specific design, where it is both open and extensible for any available semantic information.}

\subsection{Multi-view Semantic Feature Learning}
In 3D object detection, multi-view semantic features can be utilized to enhance scene representation. To promote the learning of multi-view semantic features, it is necessary to identify and learn multi-view consistent attributes. The projection-based multi-view methods \cite{zhou2020end,yang2020multi,deng2021multi} overlapped the perspective, left, and right views onto the Bird's Eye View (BEV) on the basis of spatial consistency. VISTA \cite{deng2022vista} used spatial attention to establish contextual correlations between BEV and range view, allowing for closer fusion of multi-view semantic features. DepthContrast \cite{zhang2021self} and ProposalContrast \cite{yin2022proposalcontrast} contrasted the multi-view consistency of instance features and proposals,  respectively, to enhance detection performance. These methods emphasize the importance of multi-view consistent attributes, significantly contributing to semantic feature learning. \revise{In addition, H$^2$3D R-CNN \cite{deng2021multi} projected point-wise semantic features from the perspective view into BEV in the first stage and proposed a cross-view gating fusion module in the second stage, which improved multi-view feature aggregation. However, this approach lacks direct supervision of perspective-view proposals and does not ensure multi-view proposal consistency, resulting in limited performance. To this issue, our first-stage parallel detector introduces both separate supervision for each view's proposals and unified supervision to ensure proposal-level consistency across multiple views, thereby enhancing the learning and aggregation of multi-view semantic features.}

\begin{figure*}[!htb]
\begin{center}
    \begin{tabular}{c@{\hspace{-4mm}}}
  \includegraphics[width=6.3in]{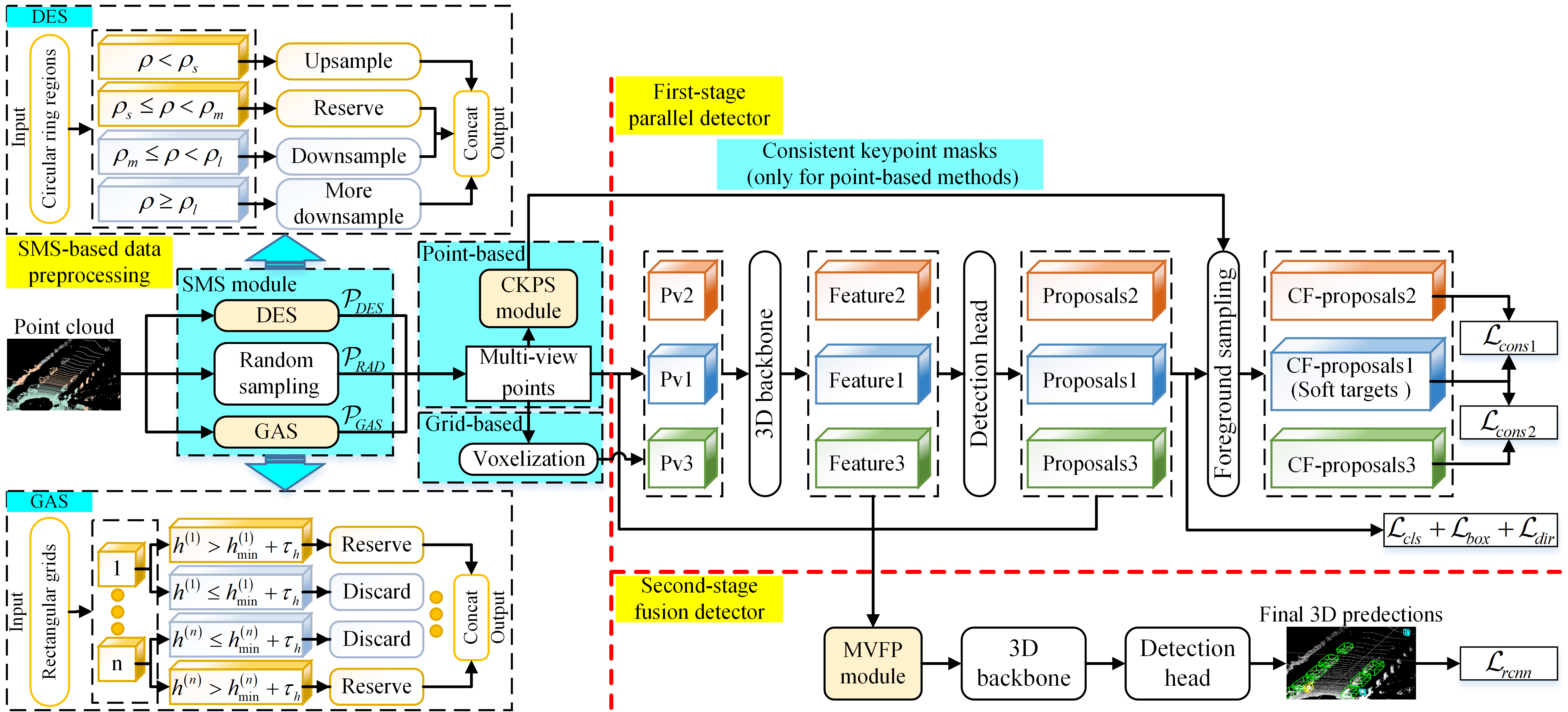}   \\
\end{tabular}
  \caption{\revise{The overall framework of our approach. In DES, points within low-density remote regions are upsampled to enhance distant objects, while the inner points of high-density near regions are downsampled to reduce point count. For these points within rectangular grids of GAS, when their height differences compared to the lowest point within each grid are lower than ${\tau _h}$, they are classified as the ground points to be discarded. The multi-view representations generated by SMS-based data preprocessing are fed into the first-stage parallel detector to extract features and predict proposals. With the assistance of the CKPS module, foreground sampling filters CF-proposals to calculate the multi-view consistency loss for feature aggregation. In the second-stage fusion detector, multi-view data from the first stage is fused through the MVFP module to obtain final 3D predictions.}}
  \label{fig:figure2}
\vskip -10pt
\end{center}
\end{figure*}

\section{The Proposed Approach}
\subsection{The Overall Framework}
In 3D object detection of point cloud, standard two-stage frameworks enable richer learning capability and higher detection accuracy compared to single-stage frameworks. However, in common two-stage 3D detectors, the grid-based and point-based sampling methods in data preprocessing often neglect the semantic information within point clouds. This results in both the partial loss of crucial object details and interference from large numbers of ground points. To solve these problems, we propose a multi-branch two-stage 3D object detection framework based on SMS and multi-view consistency constraints, as shown in Fig. \ref{fig:figure2}.

First of all, \revise{utilizing the abstract semantic information of regional density and local height,} we design a three-branch SMS module in the data preprocessing stage to generate multi-view representations (denoted as Pv1, Pv2, and Pv3) that convey enriched semantic information. Meanwhile, in order to make the subsequent foreground sampling for point-based multi-view proposals more efficient and organized, we introduce an auxiliary CKPS module that leverages multi-view consistent attributes to generate consistent keypoint masks. Then, the three-branch first-stage detector, with each branch having the same structure and parameters, performs multi-branch parallel learning to extract features and predict proposals from these multi-view representations. For training, we implement foreground sampling to obtain multi-view proposals corresponding to the Consistent Foreground (CF) points or voxels (namely CF-proposals in Fig. \ref{fig:figure2}), which are used to establish proposal-level consistency constraints for enhancing multi-view feature aggregation. Finally, the second-stage detector fuses the multi-view points, features, and proposals from the first-stage through the MVFP module to generate the final 3D predictions.

\subsection{Data Preprocessing Based on SMS Module}
In traditional single-branch sampling of data preprocessing, grid-based methods use voxelization to sample point clouds, leading to the absence of fine-grained information. In contrast, point-based methods generally employ random sampling to reduce the point count, resulting in the loss of some crucial object points. These challenges are exacerbated when detecting distant objects with fewer points, making identification more difficult. Moreover, the large number of ground points retained by both single-branch sampling methods can cause serious interference in 3D object detection. To address these issues, we propose a three-branch SMS module for data preprocessing, as depicted in Fig. \ref{fig:figure2}. \revise{This module simultaneously addresses multiple defects of single-branch sampling methods by introducing sampling branches with distinct semantic emphases.} Each branch serves a specific purpose in the sampling process: the first branch performs random sampling to produce a random point cloud (${\cal P}_{RAD}$), the second branch utilizes DES to create a density-balanced point cloud (${\cal P}_{DES}$) that enhances distant objects, and the third branch employs GAS to obtain a non-ground point cloud (${\cal P}_{GAS}$) that focuses on detection objects. \revise{The multi-view input points generated by the SMS module emphasize each branch's semantic advantages and enhance the robustness to potential shortcomings.}

\textbf{Density Equalization Sampling.} LiDAR sensors scan the traffic scene in a 360-degree range with the vehicle at the center. They collect the laser reflection points to generate the point clouds, which have the characteristics of expanding in circles and gradually reducing density from the center to the outer. When the Z-axis is not considered, the detection range of point clouds is generally a rectangular region. To explore the semantic information of regional density, we divide the detection scene into multiple circular ring regions based on the diffusion characteristics of point clouds, where the planar distance of points is denoted as ${d} \in [0,{\tau_{far}}]$. The circular ring width is expressed by $d_t$, and the divided circular ring regions are defined as ${\cal R} = \left\{ {{r_1}, \cdots ,{r_j}} \right.$ $\left. {\left| {j \in \left\{ {1, \cdots ,{n_r}} \right\}} \right.} \right\}$, where the number of circular ring regions is ${{n}_r} = {\tau_{far}}/{d_t}$. The area $S^{(r_j)}$ and point density $\rho^{(r_j)}$ of the $j_{th}$ circular ring region are determined as:
\begin{equation}
\label{1}
{S^{({r_j})}} = \mu \cdot \pi \cdot ({j^2} - {(j - 1)^2}) \cdot {d}_t^2,{\rho ^{({r_j})}} = n_p^{({r_j})}/{S^{({r_j})}}, \\
\end{equation}
where $\mu$ is the area coefficient for each region, and the point number in the $j_{th}$ circular ring region is represented by $n_p^{(r_j)}$.

According to the point density $\rho^{(r_j)}$ of each region, we define three DES thresholds: low density $\rho_{s}$, medium density $\rho_{m}$ and high density $\rho_{l}$. Points within the $j_{th} $ circular ring region are defined as ${{\cal P}^{\left( {{r_j}} \right)}} = \left\{ {p_1^{\left( {{r_j}} \right)}, \cdots ,p_k^{\left( {{r_j}} \right)}} \right.$ $\left. {\left| {k \in \left\{ {1, \cdots ,n_p^{\left( {{r_j}} \right)}} \right\}} \right.} \right\}$. The rules for obtaining points ${\cal P}_{DES}^{({r_j})}$ through upsampling or downsampling of DES are formulated as:
\begin{equation}
\label{2}
{\cal P}_{DES}^{({r_j})} = \left\{ 
\begin{array}{ll}
{\cal U}\left( {{{\cal P}^{({r_j})}}, {s_1}, {\tau _z}} \right), & \rho^{({r_j})} < \rho_s, \\
{{{\cal P}^{({r_j})}},} & \rho_s \le \rho^{({r_j})} < \rho_m, \\
{\cal D}\left( {{{\cal P}^{({r_j})}}, {s_2}} \right), & \rho_m \le \rho^{({r_j})} < \rho_l, \\
{\cal D}\left( {{{\cal P}^{({r_j})}}, {s_3}} \right), & \rho^{({r_j})} \ge \rho_l,
\end{array} 
\right.
\end{equation}
where $s_1, s_2$ and $s_3$ are sampling proportions, with $s_1=s_3>s_2$. The functions $ \cal U(\cdot)$ and $\cal D(\cdot)$ represent upsampling and downsampling operations within circular ring regions, respectively. In particular, during upsampling in low-density remote regions, only points that meet the z-focus threshold $\tau_z$ for height are upsampled, enhancing the emphasis on distant objects.

\textbf{Ground Abandonment Sampling.} In areas close to LiDAR sensors, point clouds contain a large number of dense ground points, which can comprise nearly half of the total point count. These ground points consume an abundant percentage of computing resources and cause potentially severe interference in 3D box regression. To this end, we use a planar gridding method to divide the point cloud space and filter non-ground object points according to the semantic information of local height. Here, points beyond the Z-axis detection range need to be removed in advance. The rectangular grid range satisfies $x\in[{x_s},{x_l}]$ and $y\in[{y_s},{y_l}]$, with grid sizes of the X-axis and Y-axis defined as $x_t$ and $y_t$, respectively. The number of grids along the X-axis $n_{gx}$ and Y-axis $n_{gy}$ can be calculated as:
\begin{equation}
\label{3}
{n_{gx}} = \left( {{x_l} - {x_s}} \right)/{x_t},{n_{gy}} = \left( {{y_l} - {y_s}} \right)/{y_t}.  \\
\end{equation}

The divided rectangular grids are represented as ${\cal G} = \left\{ {{g_1}, \cdots ,{g_j}} \right.$ $\left. {\left| {j \in \left\{ {1, \cdots ,{n_{gx}} \cdot {n_{gy}}} \right\}} \right.} \right\}$. We define the points within the $j_{th}$ rectangular grid as ${{\cal P}^{\left( {{g_j}} \right)}} = \left\{ {p_1^{\left( {{g_j}} \right)}, \cdots ,p_k^{\left( {{g_j}} \right)}} \right.$ $\left. {\left| {k \in \left\{ {1, \cdots ,n_p^{\left( {{g_j}} \right)}} \right\}} \right.} \right\}$. The rule for selecting points ${\cal P}_{GAS}^{(g_j)}$ by GAS can be expressed as:
\begin{equation}
\label{4}
{\cal P}_{GAS}^{\left( {{g_j}} \right)} = \left[ {{{\cal P}^{\left( {{g_j}} \right)}};h_k^{\left( {{g_j}} \right)}} \right],s.t.{\rm{ }}h_k^{\left( {{g_j}} \right)} > h_{\min }^{\left( {{g_j}} \right)} + {\tau _h},
\end{equation}
where $h_k^{(g_j)}$ is the point height in the $j_{th}$ grid, $h_{min}^{(g_j)}$ is the lowest point height in the same grid, and $\tau_h$ is the height difference threshold.

To sum up, we obtain multi-view point clouds with different semantic emphases through the SMS module. These point clouds are then transformed into various multi-view representations based on the subsequent detection network’s data requirements. For point-based detection networks, the multi-view representations are (Pv1, Pv2, Pv3) = $\left( {{{\cal P}_{RAD}},{{\cal P}_{DES}},{{\cal P}_{GAS}}} \right)$, while for grid-based detection networks, the transformed multi-view representations are (Pv1, Pv2, Pv3) = $\left(\mathbb V {\left( {{{\cal P}_{RAD}}} \right),\mathbb V \left( {{{\cal P}_{DES}}} \right),\mathbb V \left( {{{\cal P}_{GAS}}} \right)} \right)$, where $\mathbb V(\cdot)$ represents voxelization.

\subsection{CKPS Module}
\begin{figure}[!htb]
\begin{center}
    \begin{tabular}{c@{\hspace{-4mm}}}
    \hspace{-15pt}\includegraphics[width=3.5in]{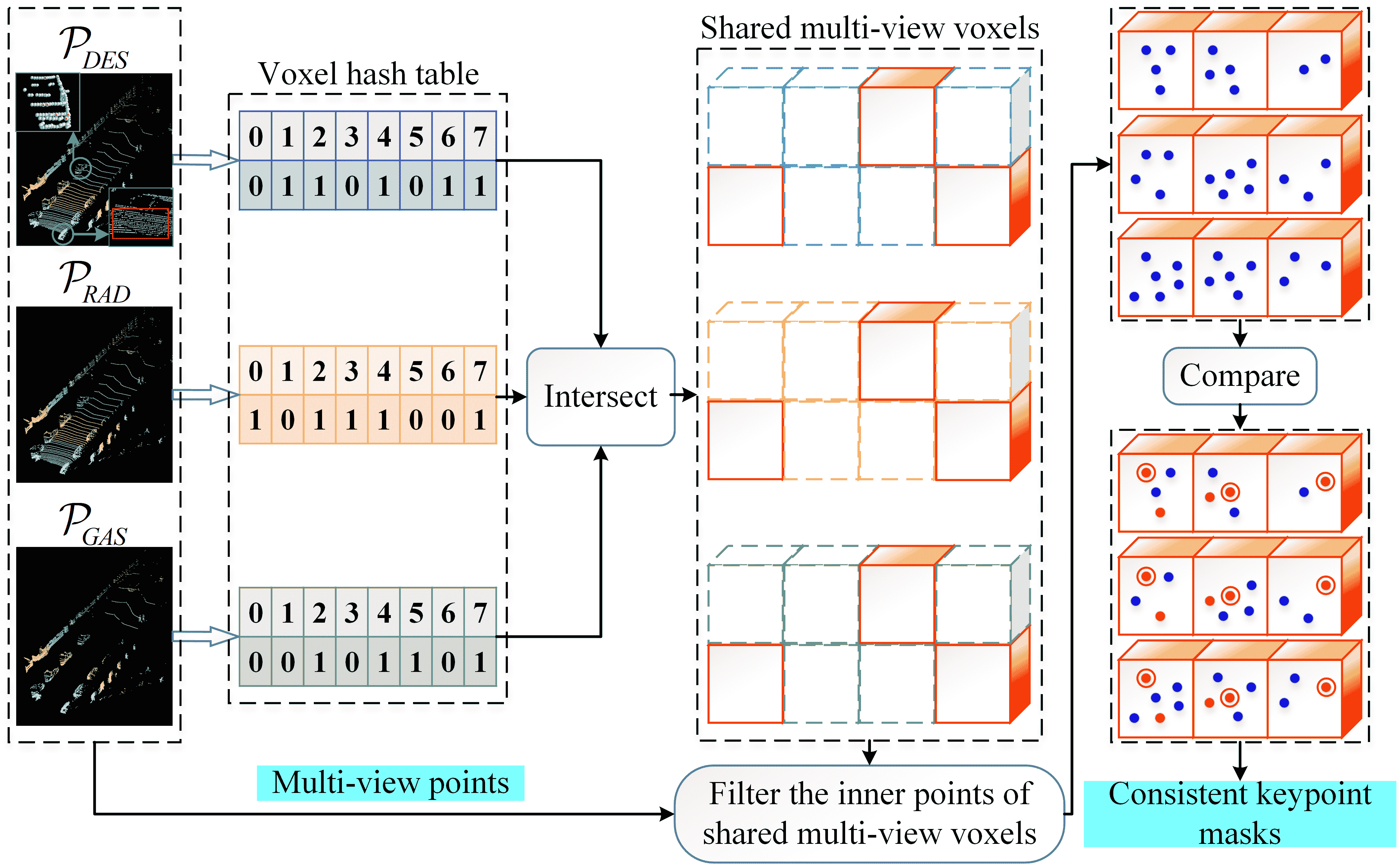}   \\
\end{tabular}
  \caption{\revise{Diagram of the CKPS module. Shared multi-view voxels (shown as red cubes) are first created by intersecting the voxel hash tables from multiple views. Next, multi-view inner points (represented by blue points) are filtered and assigned to these voxels based on spatial positions. Subsequently, consistent points (shown as red points) are selected by comparing the inner points of each voxel across multiple views. Finally, the first set of these consistent points is treated as a set of consistent keypoints (marked by red circles) to generate consistent keypoint masks with the current voxel. Best viewed in color.}}
  \label{fig:figure5}
\vskip -10pt
\end{center}
\end{figure}

To enhance multi-view feature aggregation, it is necessary to identify CF-proposals to establish the proposal-level consistency constraints across multiple views. Since point-based multi-view proposals are misaligned, dense, and overlapping, direct foreground sampling often introduces redundant calculations and reduces the effectiveness of proposal-level consistency constraints. To address this, we design an auxiliary CKPS module that uses voxel downsampling \cite{shi2020point} to process multi-view point clouds, generating spatially aligned and uniform consistent keypoint masks. Since voxel downsampling lacks an exact one-to-one correspondence between voxels and points, additional screening steps are required to obtain consistent keypoints. The diagram of CKPS module is illustrated in Fig. \ref{fig:figure5}.

Initially, multi-view point clouds are voxelized with a unified voxel size, and voxel hash tables are constructed based on voxel positions in each view, marking empty voxels as 0 and non-empty voxels as 1. By intersecting these voxel hash tables, we identify voxels with a value of 1, defining them as shared multi-view voxels ${\cal V} = \left\{ {{v_1}, \cdots ,{v_k}|k \in \left\{ {1, \cdots ,{n_v}} \right\}} \right\}$ (represented by red cubes in Fig. \ref{fig:figure5}), where $n_v$ is the number of these voxels. Next, according to spatial positions, we filter the $v_k$ inner points ${\cal P}_i^{(v_k)}$ (shown as blue points in Fig. \ref{fig:figure5}) from the $i_{th}$ view's point cloud ${{\cal P}_i} = \left\{ {{p_{ij}}|{p_{ij}} = \left( {{x_{ij}},{y_{ij}},{z_{ij}}} \right),j \in \left\{ {1, \cdots ,{n_p}} \right\}} \right\}$ using the following conditions:
\begin{equation}
\label{5}
{\cal P}_i^{\left( {{v_k}} \right)} = \left[ {{p_{ij}}} \right],s.t.{\rm{ }}\left\{ {\begin{array}{*{20}{c}}
{s{x^{\left( {{v_k}} \right)}} \le {x_{ij}} \le l{x^{\left( {{v_k}} \right)}}},\\
{s{y^{\left( {{v_k}} \right)}} \le {y_{ij}} \le l{y^{\left( {{v_k}} \right)}}},\\
{s{z^{\left( {{v_k}} \right)}} \le {z_{ij}} \le l{z^{\left( {{v_k}} \right)}}},
\end{array}} \right.i \in \left\{ {1,2,3} \right\},  \\
\end{equation}
where $n_p$ is the fixed number of points in a single view, and $(s{x^{({v_k})}},l{x^{ ({v_k}) }},s{y^{ ({v_k}) }},l{y^{ ({v_k}) }},s{z^{ ({v_k}) }},l{z^{ ({v_k}) }} )$ are the common boundary values of $v_k$ derived from voxel sizes and positions. We then compare coordinates and reflectivity between views for the inner points $ {\cal P}_i^{(v_k)}$ of each shared multi-view voxel $v_k$. Specifically, the $v_k$ inner points of $i_{th}$ view are ${\cal P}_i^{\left( {{v_k}} \right)} = \left\{ {{p_{im}}\left| {{p_{im}} = \left( {{x_{im}},{y_{im}},{z_{im}},{r_{im}}} \right),m \in \left\{ {1, \cdots ,n_{pi}^{\left( {{v_k}} \right)}} \right\}} \right.} \right\}$, and in the first view, they are ${\cal P}_1^{\left( {{v_k}} \right)} = \left\{ {{p_{1j}}\left| {{p_{1j}} = \left( {{x_{1j}},{y_{1j}},{z_{1j}},{r_{1j}}} \right),j \in \left\{ {1, \cdots ,n_{p1}^{\left( {{v_k}} \right)}} \right\}} \right.} \right\}$, where $(x_{im}, y_{im}, z_{im}, r_{im})$ and $(x_{1j}, y_{1j}, z_{1j}, r_{1j})$ represent the coordinates and reflectivity of the $v_k$ inner points in different views. We formulate the rules of obtaining consistent keypoints ${\cal P}_c^{(v_k)}$ by comparing ${\cal P}_i^{(v_k)}$ and ${\cal P}_1^{(v_k)}$ as:

\begin{equation}
\label{6}
\begin{split}
& {\cal P}_{c}^{\left( {{v_k}} \right)} = \left[ {{p_{c}^{\left( {{v_k}} \right)}}|{p_{c}^{\left( {{v_k}} \right)}} \in [{\cal P}_i^{\left( {{v_k}} \right)} \cup {\cal P}_1^{\left( {{v_k}} \right)}] }\right],   \\
& s.t. \vert\vert{{p_{im}} - {p_{1j}}}\vert\vert_{\infty} < {\tau _v} , i \in \left\{{2,3}\right\} ,
\end{split}
\end{equation}
where $\tau_v$ is the minimum threshold of numerical comparison. As shown in Fig. \ref{fig:figure5}, the red points within each $v_k$ represent potential consistent points that satisfy these conditions. The first set of consistent points is defined as consistent keypoints, marked by red circles in Fig. \ref{fig:figure5}. Finally, we use the indices of these consistent keypoints ${{\cal P}_c} = \left\{ {{\cal P}_c^{({v_1})}, \cdots ,{\cal P}_c^{({v_k})}|k \in \left\{ {1, \cdots ,{n_v}} \right\}} \right\}$ as consistent keypoint masks for foreground sampling of point-based multi-view proposals. In summary, the CKPS module selects consistent keypoints characterized by spatial alignment, uniformity, limited quantity, and high quality, effectively enhancing the computational efficiency and constraint effectiveness of the subsequent point-based multi-view consistency loss.

\subsection{Multi-view Learning Strategy}
For the multi-view point clouds with distinct semantic emphases generated by the SMS module, this paper introduces strategies for first-stage parallel training and second-stage fusion learning. The main methods can be summarized in following three aspects.

\textbf{Multi-view Foreground Sampling.} Before using the multi-view consistency loss, foreground sampling aims to filter CF-proposals in the first-stage detector. In grid-based networks, the multi-view proposals generated by the first-stage detector are represented as ${\cal P}{\cal P}_i^{(bev)} = \left\{ {pp_i^{(be{v_1})}, \cdots ,pp_i^{(be{v_j})}\left| {j \in \left\{ {1, \cdots ,{n_{bev}}} \right\}} \right.} \right\}$, where $n_{bev}$ is the number of BEV grids in the current space. Since grid-based multi-view proposals are spatially aligned on BEV, the CF-proposals ${\cal CPP}_i^{(bev_{fg})}$ obtained through foreground sampling can be expressed as follows:
\begin{equation}
\label{7}
{\cal C}{\cal P}{\cal P}_i^{(be{v_{fg}})} = {\cal M}_i^{(bev)}\left( {{\cal P}{\cal P}_i^{(bev)}} \right),i \in \left\{ {1,2,3} \right\},
\end{equation}
where ${\cal M}_i^{(bev)}\left(  \cdot  \right)$ denotes the process of filtering proposals corresponding to the foreground BEV grids in GT.

In point-based networks, multi-view proposals are ${\cal P}{\cal P}_i^{(p)} = \left\{ {pp_i^{({p_1})}, \cdots ,pp_i^{({p_j})}\left| {j \in \left\{ {1, \cdots ,{n_p}} \right\}} \right.} \right\}$. Due to the analysis and design of the CKPS module, we implement foreground sampling through the following formula to obtain CF-proposals ${\cal C}{\cal P}{\cal P}_i^{({p_{fg}})}$:
\begin{equation}
\label{8}
{\cal C}{\cal P}{\cal P}_i^{({p_{fg}})} = {\cal M}_i^{(p)}\left( {{\cal C}_i^{\left( p \right)}\left( {{\cal P}{\cal P}_i^{(p)}} \right)} \right),i \in \left\{ {1,2,3} \right\},
\end{equation}
where ${\cal C}_i^{\left( p \right)}\left(  \cdot  \right)$ represents proposal sampling based on the consistent keypoint masks from the CKPS module, and ${\cal M}_i^{(p)}\left(  \cdot  \right)$ is the operation for screening proposals corresponding to the foreground points in GT.

\textbf{Multi-view Consistency Loss.} SE-SSD \cite{zheng2021se} uses both the soft target from teacher network and the hard target from GT to jointly supervise a student network. The CF-proposals from the pre-trained teacher network, which exhibit higher entropy than GT, provide stronger proposal-level consistency constraints by comparing their consistency with the CF-proposals of the student network. Similarly, in this paper, we use the CF-proposals ${\cal C}{\cal P}{\cal P}_1^{({p_{fg}},be{v_{fg}})}$ from the first view as soft targets for proposal-level consistency comparison with the CF-proposals ${\cal CPP}_i^{({p_{fg}},be{v_{fg}})}$ from other views. We define the number of foreground points or BEV grids as $N_{fg}$, and the multi-view consistency loss for bounding boxes ${\cal L}_{box}^{{\rm{ }}c}$ can be formulated as:
\begin{equation}
\label{9}
{\cal L}_{box}^{{\rm{ }}c} = \frac{1}{{{N_{fg}}}}\sum\limits_{j = 1}^{{N_{fg}}} {\frac{1}{7}} \sum\limits_b {{\cal L}_{\delta _b^{\left( j \right)}}^{{\rm{ }}c}} ,\delta _b^{\left( j \right)} = \left| {b_i^{\left( j \right)} - b_1^{\left( j \right)}} \right|,i \in \left\{2,3 \right\}, \\
\end{equation}
where $ {b_1^{\left( j \right)}}$ and ${b_i^{\left( j \right)}}$ are the bounding boxes of CF-proposals in the respective views. ${{\cal L}_{\delta _b^{\left( j \right)}}^{{\rm{ }}c}}$ denotes the Smooth-L1 loss of $\delta _b^{\left( j \right)}$. Then, we define the multi-view consistency loss for classification scores ${\cal L}_{cls}^{{\rm{ }}c}$ as
\begin{equation}
\label{10}
{\cal L}_{cls}^{{\rm{ }}c} = \frac{1}{{{N_{fg}}}}\sum\limits_{j = 1}^{{N_{fg}}} {{\cal L}_{\delta _c^{\left( j \right)}}^{{\rm{ }}c}} ,\delta _c^{\left( j \right)} = \left| {\sigma \left( {c_i^{\left( j \right)}} \right) - \sigma \left( {c_1^{\left( j \right)}} \right)} \right|,i \in \left\{2,3 \right\}, \\
\end{equation}
where ${\sigma \left( {c_1^{\left( j \right)}} \right)}$ and ${\sigma \left( {c_i^{\left( j \right)}} \right)}$ are the sigmoid-activated classification scores of CF-proposals in the respective views. ${{\cal L}_{\delta _c^{\left( j \right)}}^{{\rm{ }}c}}$ denotes the Focal Loss of $\delta _c^{\left( j \right)}$. In the aforementioned multi-view consistency losses, we minimize differences in the classification scores and bounding boxes of the CF-proposals across multiple views.
To mitigate the linear variation of the multi-view consistency loss with the number of views $N_{mv}$, we formulate the multi-view consistency weight $\gamma _{mv}^c$ according to an inverse proportion principle. The overall multi-view consistency loss ${{\cal L}_{cons}}$ is defined as:
\begin{equation}
\label{11}
{{\cal L}_{cons}} = \gamma _{mv}^c\left( {{\cal L}_{cls}^{{\rm{ }}c} + {\cal L}_{box}^{{\rm{ }}c}} \right),\gamma _{mv}^c = \frac{1}{{{N_{mv}} - 1}}. \\
\end{equation}

\begin{figure}[!htb]
\begin{center}
    \begin{tabular}{c@{\hspace{-4mm}}}
    \hspace{-15pt}\includegraphics[width=3in]{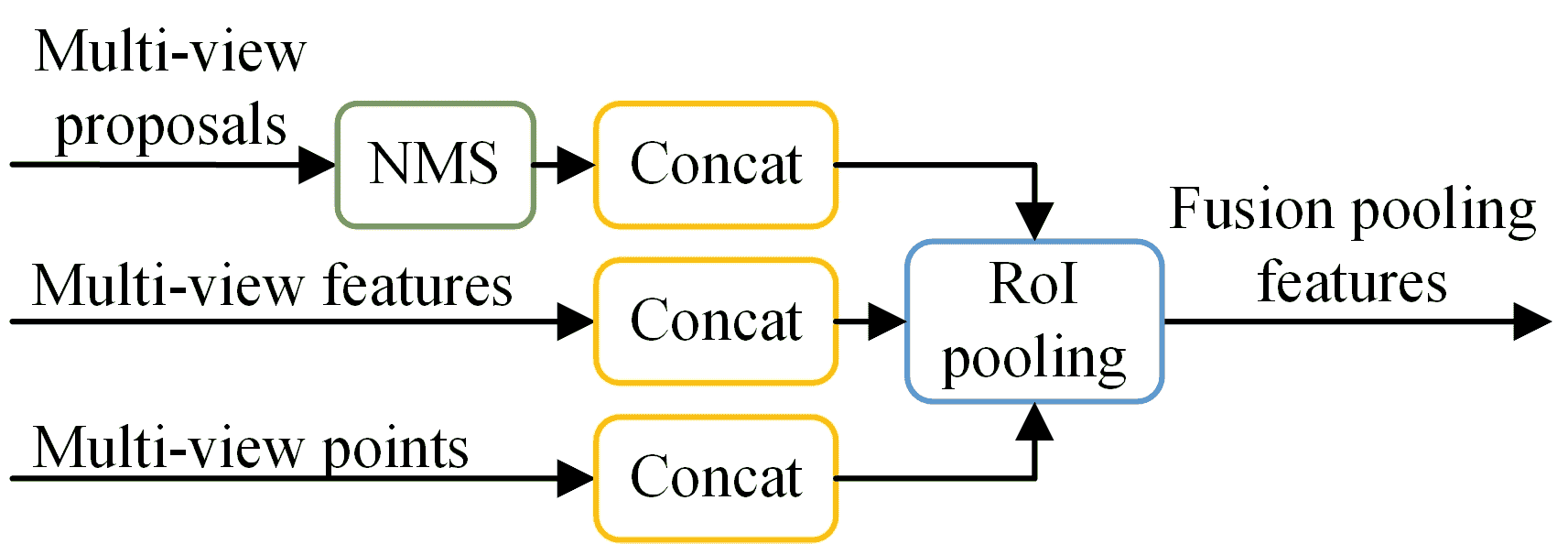}   \\
\end{tabular}
  \caption{Schematic of the MVFP module. Multi-view data is concatenated and then fed into the RoI pooling operation to generate fusion pooling features.}
  \label{fig:figure6}
\vskip -10pt
\end{center}
\end{figure}

\textbf{MVFP Module}. In existing two-stage 3D object detection networks \cite{shi2019pointrcnn,shi2020pv}, the second-stage detector usually performs Non-Maximum Suppression (NMS) sampling on proposals to obtain a small set of high-quality proposals. Subsequently, the obtained proposals are resampled to generate RoIs according to specific IoU rules. Within the RoIs, keypoints and their features are aggregated through RoI pooling to obtain pooling features. For multi-view fusion, we design the MVFP module to perform these operations, as shown in Fig. \ref{fig:figure6}. Let ${\cal F}_i$ represent the features associated with the multi-view points ${\cal P}_i$, and ${{\cal P}{\cal P}_i^{\left( {p,bev} \right)}}$ denote the multi-view proposals generated by the first-stage detector. Then fusion pooling features ${{\cal F}_p}$ generated by the MVFP module can be expressed as follows:
\begin{equation}
\label{12}
\mathcal{F}_p = \mathbb{P}\left(\mathbb{C}(\mathcal{P}_i), \mathbb{C}\left(\mathcal{F}_i\right), \mathbb{C}\left(\mathbb{N}\left(\mathcal{P}\mathcal{P}_i^{({p, bev})}\right)\right)\right),i \in \left\{1,2,3 \right\}, \\
\end{equation}
where $\mathbb{N} (\cdot ) $ represents NMS sampling of proposals, $\mathbb{C} (\cdot ) $ denotes data concatenation, and $\mathbb{P} (\cdot )$ is the operation for extracting pooling features within RoIs. The MVFP module effectively integrates various features from multiple views to refine predictions.

\subsection{Loss Function}
Referring to the multi-view consistency weight $\gamma_{mv}^c$, we employ multi-view weights $\gamma_{mv}$ for the detection losses in the first-stage parallel detector. The overall loss function is defined as:
\begin{equation}
\label{13}
\begin{split}
& \mathcal{L} = \mathcal{L}_{cons} + \gamma_{mv} (\lambda_1 \mathcal{L}_{cls} + \lambda_2 \mathcal{L}_{box} + \lambda_3 \mathcal{L}_{dir}) + \mathcal{L}_{rcnn},   \\
& \gamma_{mv}^c = \frac{1}{N_{mv}},
\end{split}
\end{equation}
where $\mathcal{L}_{cls}, \mathcal{L}_{box},$ and $\mathcal{L}_{dir}$ are the classification, bounding box and direction losses for the first-stage detector, respectively; $\lambda_1, \lambda_2$ and $\lambda_3$ are their corresponding weight coefficients. $\mathcal{L}_{cons}$ is the multi-view consistency loss of the first-stage detector; and $\mathcal{L}_{rcnn}$ is the comprehensive loss for the second-stage detector. The weight coefficients for $\mathcal{L}_{cons}$ and $\mathcal{L}_{rcnn}$ are set to one.

\section{Experiments}
\revise{We evaluate our semantic-aware multi-branch framework on the popular KITTI dataset and the challenging Waymo Open Dataset (WOD).} First, Section \hyperref[section:4A]{\ref{section:4A}} outlines the implementation details. Then, the proposed approach is compared with advanced 3D object detection methods in Section \hyperref[section:4B]{\ref{section:4B}}. Finally,  Section \hyperref[section:4C]{\ref{section:4C}} presents extensive ablation studies to analyze the effectiveness of the designed modules.

\begin{table*}[!t]
\centering
\caption{\revise{The 3D object detection performance on the KITTI test split, which includes AP and mAP of 40 recall positions. Bolded and underlined values are the best and suboptimal performances for all methods, respectively. *: The reimplementation result of a public code by training a single model for three classes.}}
\label{table:table1}
\begin{tabular}{c|c|cccc|cccc|cccc}
\hline
\multirow{2}{*}{Method} &  \multirow{2}{*}{Type}  & \multicolumn{4}{c|}{Car 3D} & \multicolumn{4}{c|}{Ped. 3D}  & \multicolumn{4}{c}{Cyc. 3D}             \\
& & {Easy} & {Mod.} & {Hard} & {mAP}
& {Easy} & {Mod.} & {Hard} & {mAP}
& {Easy} & {Mod.} & {Hard} & {mAP} \\
\hline
\hline
PointRCNN \cite{shi2019pointrcnn} & point-based & 86.96 & 75.64 & 70.7 & 77.77 & 47.98 &39.37 & 36.01 & 41.12 & 74.96 & 58.82 & 52.53 & 62.10 \\
IA-SSD \cite{zhang2022not} & point-based & 88.87 & 80.32 & 75.10 & 81.43 & 49.01 & 41.20 & 38.03 & 42.75 & 80.78 & 66.01 & 58.12 & 68.30 \\
FARP-Net \cite{xie2023farp} & point-based & 88.36 & 81.53          &\textbf{78.98} & 82.96 & - & - & - & - & - & - & - & - \\
\hline
SMS-PointRCNN(ours) & point-based & 86.12 & 78.71 & 74.17 & 79.67 & 50.3 & 41.18 & 37.41 & 43.06 & 75.92 & 61.98 & 54.62 & 64.17 \\
\hline
S-AT GCN \cite{wang2021s} & grid-based & 83.20 & 76.04 & 71.17 & 76.80 & 44.63 & 37.37 & 34.92 & 38.97 & 75.24 & 61.70 & 55.32 & 64.09  \\
Part-A\textsuperscript{2}{\cite{shi2020points}} & grid-based & 85.94 & 77.86 & 72.00 & 78.60 & \textbf{54.49} & \underline{44.50} & \textbf{42.36} & \textbf{47.12} & 78.58 & 62.73 & 57.74 & 66.35\\
HVPR \cite{noh2021hvpr} & grid-based & 86.38 & 77.92 & 73.04 & 79.11 & - & - & - & - & - & - & - & - \\
Faraway-Frustum \cite{zhang2021faraway} & grid-based & 87.45 & 79.05 & 76.14 & 80.88 & 46.33 & 38.58 & 35.71 & 40.21 & 77.36 & 62.00 & 55.40 & 64.92 \\
H\textsuperscript{2}3D R-CNN \cite{deng2021multi} & grid-based & \underline{90.43} & 81.55 & \underline{77.22} & \underline{83.07} & \underline{52.75} & \textbf{45.26} & \underline{41.56} & \underline{46.52} & 78.67 & 62.74 & 55.78 & 65.73 \\
Voxel-RCNN \cite{deng2021voxel}* & grid-based & 88.09 & 80.99 & 76.50 & 81.86 & 47.91 & 40.57 & 38.21 & 42.23 & 76.42 & 62.01 & 55.94 & 64.79 \\
SVGA-Net \cite{he2022svga} & grid-based & 87.33 & 80.47 & 75.91 & 81.24 & 48.48 & 40.39 & 37.92 & 42.26 & 78.58 & 62.28 & 54.88 & 65.25 \\
BADet \cite{qian2022badet} & grid-based & 89.28 & \underline{81.61} & 76.58 & 82.49 & - & - & - & - & - & -  & - & -  \\
X-view \cite{xie2023x} & grid-based & 89.21 & 81.35 & 76.87 & 82.48 & - & - & - & - & - & -  & - & -  \\
PV-RCNN \cite{shi2020pv} & grid-based & 90.25 & 81.43 & 76.82 & 82.83 & 52.17 & 43.29 & 40.29 & 45.25 & 78.60 & 63.71 & 57.65 & 66.65 \\
PV-RCNN \cite{shi2020pv}* & grid-based & 90.39 & 81.58 & 77.02 & 83.00 & 45.97 & 39.33 & 37.03 & 40.78 & 80.35 & 64.75 & 57.73 & 67.61 \\
PV-RCNN++ \cite{shi2023pv}* & grid-based & 87.99 & 81.37 & 76.56 & 81.97 & 50.30 & 42.00 & 39.80 & 44.03 & \underline{84.49} & \underline{67.65} & \textbf{61.03} & \underline{71.06} \\
\hline
SMS-PVRCNN(ours) & grid-based & \textbf{90.85} & \textbf{81.72} & 77.12 & \textbf{83.23} & 46.68 & 40.55 & 38.36 & 41.86 & 81.58 & 66.12 & 59.67 & 69.12 \\
SMS-PVRCNN++(ours) & grid-based & 88.55 & 81.45 & 76.55 & 82.18 & 51.01 & 42.80 & 40.61 & 44.81 & \textbf{85.71} & \textbf{68.05} & \underline{59.96} & \textbf{71.24} \\
\hline
\end{tabular}
\end{table*}

\subsection{Implementation Details}
\label{section:4A}
\subsubsection{Datasets and Hardware}
For evaluating the proposed method on the KITTI validation split, we divide the training samples into two subsets: a training split with 3712 samples and a validation split with 3769 samples. The performance of our multi-branch two-stage 3D object detection framework is evaluated on the test split, where we use all training samples for training process. Our KITTI models are trained for 80 epochs with a batch size of 2 on an RTX 3090 GPU.

\revise{To prove the generalizability of the proposed method on large datasets, we conduct experiments on the WOD using 20\% of the training sequences (about 32k frames) and the entire validation sequences. Objects in each WOD category are divided into two difficulty levels: LEVEL 1 (L1) for objects with more than 5 points and LEVEL 2 (L2) for objects have 1\~{}5 points. Our WOD models are trained for 30 epochs with a batch size of 8 on four RTX 4090 GPUs.}

\subsubsection{Training Parameters For KITTI}
We set the fixed number of sampling points $n_p$ to 16384, and the distance threshold for distant points $\tau_{far}$ to 40. For the DES component of the SMS module, the point cloud space is divided into circular ring regions with a width of $d_t=5$ and an area coefficient of $\mu=0.5$. DES adjusts the point percentages within these regions based on the semantic information of regional density. The low, medium, and high density thresholds are set to $\rho_s=5$, $\rho_m=8$, and $\rho_l=15$, respectively, with sampling proportions $s_1=s_3=0.15$ and $s_2=0.1$. The z-focus threshold for upsampling is $\tau_z\in[-1.5,0.5]$ in low-density remote regions. In the SMS module, GAS employs a planar gridding method to divide the point cloud space into rectangular grids with the sizes of $x_t=5$ in X-axis and $y_t=10$ in Y-axis. The rectangular grid range satisfies $x\in[0,40]$ and $y\in[-35,35]$. In each rectangular grid, GAS removes points if their height difference from the lowest point is below a threshold of $\tau_h=0.2$. The CKPS module sets a minimum threshold of numerical comparison $\tau_v$ to 0.001 to generate consistent keypoint masks. 

In the first-stage detector, we perform foreground sampling for multi-view proposals and introduce the multi-view consistency loss into the overall loss function, with the number of views $ N_{mv} $ set to 3. The weighting coefficients $(\lambda_1,\lambda_2,\lambda_3 )$ for first-stage losses are set to (1, 1, 0) for PointRCNN, (1, 2, 0.2) for PV-RCNN \revise{and (1, 2, 0) for PV-RCNN++}.

\subsubsection{Training Parameters For WOD}
\revise{According to the characteristics of WOD, we modify certain parameters as follows. The fixed number of sampling points $n_p$ is increased to 180000, and the distance threshold for distant points $\tau_{far}$ is set to 55. For DES, the circular ring regions use an area coefficient $\mu=1$, with the low, medium, and high density thresholds set to $\rho_s=12$, $\rho_m=20$, and $\rho_l=36$. The z-focus threshold for upsampling is adjusted to $\tau_z\in[-1,2]$. In GAS, the rectangular grids have the sizes of $x_t=10$ in X-axis and $y_t=10$ in Y-axis. The rectangular grid range satisfies $x\in[-45,45]$ and $y\in[-45,45]$, with a height difference threshold of $\tau_h=0.5$.}

\subsubsection{Evaluation Metrics}
As to 3D detection performance on the validation and test splits of the KITTI dataset, the primary evaluation metric is the average precision (AP) of 40 recall positions, with 3D IoU thresholds of 0.5 for pedestrians, cyclists, and 0.7 for cars. The mean Average Precision (mAP) across three difficult levels for each category is used to facilitate performance comparison. \revise{For the WOD, we employ both AP and the average precision weighted by heading (APH) for evaluation.}

\subsection{Experimental Results}
\label{section:4B}

\begin{table}[!t]
\centering
\caption{\revise{The performance comparison between baseline networks and our methods on the KITTI test split, which includes the mAP of 40 recall positions. Bolded and underlined values are the best and suboptimal performances for all methods, respectively. *: The reimplementation result of a public code by training a single model for three classes.}}
\label{table:table2}
\begin{tabular}{c|c|c|c|c}
\hline
Method & Type & Car 3D & Ped. 3D & Cyc. 3D \\
\hline
\hline
PointRCNN \cite{shi2019pointrcnn} & point-based & 77.77 & 41.12 & 62.10  \\
SMS-PointRCNN(ours) & point-based & 79.67 & 43.06 & 64.17 \\
\hline
PV-RCNN \cite{shi2020pv} & grid-based & 82.83 & \textbf{45.25} & 66.65 \\
PV-RCNN \cite{shi2020pv}* & grid-based & \underline{83.00} & 40.78 & 67.61 \\
PV-RCNN++ \cite{shi2023pv}* & grid-based & 81.97 & 44.03 & \underline{71.06} \\
SMS-PVRCNN(ours) & grid-based & \textbf{83.23} & 41.86 & 69.12  \\
SMS-PVRCNN++(ours) & grid-based & 82.18 & \underline{44.81} & \textbf{71.24}  \\
\hline
\end{tabular}
\end{table}

\revise{As shown in Table \ref{table:table1} and Table \ref{table:table2}, we report the 3D results of our approach alongside several state-of-the-art methods on the KITTI test split. Compared to previous advanced methods, SMS-PVRCNN achieves the highest mAP of 83.23 on cars, while SMS-PVRCNN++ also attains the best mAP of 71.24 on cyclists. Using the PointRCNN backbone, SMS-PointRCNN outperforms the baseline across all classes by a large margin, increasing mAP by 1.9, 1.94, and 2.07 in the car, pedestrian, and cyclist categories, respectively. Moreover, when considering the reimplementation results of the PV-RCNN baseline, SMS-PVRCNN can improve mAP by 0.23, 1.08, and 1.51 for cars, pedestrians, and cyclists. Similarly, for the PV-RCNN++ baseline, SMS-PVRCNN++ increases mAP by 0.21, 0.78, and 0.18 across the three object categories. These results indicate that our approach achieves consistent performance improvements for both point-based and grid-based backbones.} 


\revise{Table \ref{table:table3} shows our 3D results for cars on the validation split, where SMS-PVRCNN achieves the highest AP and mAP values across all levels for cars. Compared with the PointRCNN baseline, SMS-PointRCNN increases the car's mAP by 2.99. In addition, according to reimplementation results, SMS-PVRCNN and SMS-PVRCNN++ improve the car's mAPs of their respective baselines by 0.64 and 0.87. These results prove the consistent effectiveness of our method. Furthermore, our approach achieves greater performance gains for baselines with lower capabilities, such as the PointRCNN network.}

\revise{To further compare 3D detection performance on large datasets, we report 3D results on the WOD validation split in Table \ref{table:waymo}. For the PV-RCNN baseline, SMS-PVRCNN enhances AP/APH across all categories and difficulty levels, showing excellent performance improvements in the pedestrian and cyclist categories, which is consistent with our results on the KITTI dataset. For the PV-RCNN++ baseline, SMS-PVRCNN++ slightly improves AP/APH for L1 pedestrians and all levels for cars and cyclists.} 

\begin{table}[!t]
\centering
\caption{\revise{The 3D object detection Performance on the KITTI validation split for the car class, which includes AP of 40 recall positions. Bolded and underlined values are the best and suboptimal performances for all methods, respectively. *: The reimplementation result of a public code by training a single model for three classes.}}
\label{table:table3}
\setlength\tabcolsep{5pt}
\begin{tabular}{c|c|cccc}
\hline
\multirow{2}{*}{Method} & \multirow{2}{*}{Type} & \multicolumn{4}{c}{Car 3D}                               \\
&  & Easy & Mod. & Hard & mAP    \\
\hline
\hline
PointRCNN \cite{shi2019pointrcnn} & point-based & 88.88 & 78.63 & 77.38  & 81.63           \\
SMS-PointRCNN(ours) & point-based & 91.5 & 82.33 & 80.03 & 84.62 \\
\hline
PV-RCNN \cite{shi2020pv} & grid-based & \underline{92.57} & \underline{84.83} & \underline{82.69 } & \underline{86.7} \\
PV-RCNN \cite{shi2020pv}* & grid-based & 91.97 & 84.72 & 82.32 & 86.34  \\
PV-RCNN++ \cite{shi2023pv}* & grid-based & 91.38 & 83.02 & 82.03 & 85.48  \\
SMS-PVRCNN(ours) & grid-based & \textbf{92.81} & \textbf{85.36} & \textbf{82.76} & \textbf{86.98}  \\
SMS-PVRCNN++(ours) & grid-based & 91.90 & 84.84 & 82.32 & 86.35  \\
\hline
\end{tabular}
\end{table}

\begin{table*}[!t]
\centering
\caption{\revise{The 3D object detection performance on the entire validation split of WOD. All results are obtained by training on the 20\% training split (about 32k frames), and the results of each cell are AP/APH. Bolded and underlined values are the best and suboptimal performances for all methods, respectively. *: The reimplementation result of a public code by training a single model for three classes.}}
\label{table:waymo}
\begin{tabular}{c|c|c|c|c|c|c|c}
\hline
Method & Type & Vehicle(L1) & Vehicle(L2) & Pedestrian(L1) & Pedestrian(L2) & Cyclist(L1) & Cyclist(L2) \\
\hline
\hline
PV-RCNN \cite{shi2020pv}* & grid-based & 75.41/74.74 & 67.44/66.80 & 71.98/61.24 & 63.70/53.95 & 65.88/64.25 & 63.39/61.82 \\
PV-RCNN++ \cite{shi2023pv}* & grid-based & \underline{77.11/76.63} & \underline{68.70/68.26} & \underline{77.89/71.26} & \textbf{69.73/63.57} & \underline{71.60/70.51} & \underline{69.25/68.13} \\
SMS-PVRCNN(ours) & grid-based & 76.62/76.00 & 68.12/67.55 & 74.15/64.05 & 65.13/56.06 & 68.65/67.05 & 66.14/64.59 \\
SMS-PVRCNN++(ours) & grid-based & \textbf{77.67/77.20} & \textbf{69.23/68.80} & \textbf{78.03/71.52} & \underline{69.39/63.40} & \textbf{72.52/71.44} & \textbf{69.86/68.82} \\
\hline
\end{tabular}
\end{table*}

\subsection {Ablation Study}
\label{section:4C}
In this work, extensive ablation studies are conducted to validate the effectiveness of a multi-branch two-stage 3D object detection framework based on SMS and multi-view consistency constraints. The main evaluation metric is mAP.

\subsubsection{Experimental Analysis for SMS-PointRCNN}
SMS-PointRCNN, using a backbone based on PointRCNN, conducts experiments on the KITTI validation split. The following experimental results elaborate on the roles of specific parameters and modules in this paper.

\textbf{Different Modules of Multi-branch Two-stage Framework}. In data preprocessing, the proposed three-branch SMS module generates multi-view point clouds enriched with semantic features. Pv1 is the output of random sampling branch, while Pv2 and Pv3 are produced by the semantic-aware sampling branches DES and GAS, respectively. With consistent keypoint masks generated by the CKPS module, the first-stage detector calculates the multi-view consistency loss ${{\cal L}_{cons}}$ enforcing proposal-level consistency constraints across multiple views. The second-stage detector then performs fusion learning of multi-view features through the MVFP module. Building upon CKPS+${{\cal L}_{cons}}$ and the MVFP module, we evaluate the detection performance of different combinations for multi-view point clouds, as shown in Table \ref{table:table4}, where the first line indicates the baseline performance. When all views are used, the 3D mAP values for pedestrians and cars are highest. We studied the impact of CKPS+${{\cal L}_{cons}}$ and the MVFP module on detection performance using all multi-view point clouds. As shown in Table \ref{table:table5}, applying CKPS+${{\cal L}_{cons}}$ and the MVFP module increases the 3D mAPs for pedestrians and cars by 3.03 and 0.97, respectively.

\begin{table}[!h]
\centering
\caption{The 3D detection performance of different combinations for multi-view point clouds, with mAP of 40 recall positions on the KITTI validation split. Bolded and underlined values are the best and suboptimal performances for all methods, respectively.}
\label{table:table4}
\begin{tabular}{c|c|c|c|c|cc}
\hline
CKPS+${{\cal L}_{cons}}$ & MVFP & Pv1 & Pv2 & Pv3 & Ped. 3D & Car 3D  \\
\hline
\hline
${\rm{ \times }}$    & ${\rm{ \times }}$    & $\surd $           & ${\rm{ \times }}$     & ${\rm{ \times }}$   & 54.43            & 81.88            \\
$\surd $             & $\surd $             & $\surd $           & $\surd $              & ${\rm{ \times }}$   & 58.80            & \underline{84.46}            \\
$\surd $             & $\surd $             &$\surd $            & ${\rm{ \times }}$     & $\surd $            & \underline{60.41}            & 83.77            \\
$\surd $              & $\surd $            & $\surd $           & $\surd $              & $\surd $            & \textbf{63.57}   & \textbf{84.62}   \\
\hline
\end{tabular}
\end{table}

\begin{table}[!h]
\centering
\caption{The effectiveness of the CKPS+${{\cal L}_{cons}}$ and MVFP module, with mAP of 40 recall positions on the KITTI validation split. Bolded and underlined values are the best and suboptimal performances for all methods, respectively.}
\label{table:table5}
\begin{tabular}{c|c|c|cc}
\hline
Pv1+Pv2+Pv3 & CKPS+${{\cal L}_{cons}}$ & MVFP & Ped. 3D & Car 3D  \\
\hline
\hline
$\surd $          & ${\rm{ \times }}$      & ${\rm{ \times }}$    & 60.54            & 83.65            \\
$\surd $          & ${\rm{ \times }}$      & $\surd $             & \underline{61.54}            & \textbf{84.82}   \\
$\surd $          & $\surd $               & ${\rm{ \times }}$    & 61.10             & 84.45            \\
$\surd $          & $\surd $               & $\surd $             & \textbf{63.57}   & \underline{84.62}            \\
\hline
\end{tabular}
\end{table}

\textbf{Different Parameters of SMS module}. The SMS module includes two primary semantic-aware sampling branches: DES and GAS. The DES branch uses a circular ring region width $d_t$, and a z-focus threshold $\tau_z$. Meanwhile, the GAS branch has a Y-axis grid size $y_t$ and a height difference threshold $\tau_h$. Various values are assigned to these parameters to determine an optimal configuration. First, with $d_t=5$ and $\tau_z \in [-3,1]\ $ (matching the preset Z-axis detection range for KITTI), we validate the two GAS parameters. The results are shown in Table \ref{table:table6}, where it is observed that when $y_t=10$ and $\tau_h=0.2$, the 3D mAPs for pedestrians and cars reach their highest values. With these optimal GAS settings, we then validate the two DES parameters,  as shown in Table \ref{table:table7}. It is found that when $d_t=5$ and $\tau_z \in [-1.5,0.5]\ $, the 3D mAPs for both pedestrians and cars improve slightly.

\begin{table}[!h]
\centering
\caption{The 3D detection performance for different settings of GAS on the KITTI validation split, along with fixed parameters of DES. The evaluation metric is mAP of 40 recall positions. Bolded and underlined values are the best and suboptimal performances for all methods, respectively.}
\label{table:table6}
\begin{tabular}{cc|cc|c|c|cc}
\hline
\multicolumn{2}{c|}{$y_t$}  & \multicolumn{2}{c|}{$\tau_t$}       & $d_t$             & $\tau_z$           & \multirow{2}{*}{Ped. 3D} & \multirow{2}{*}{Car 3D}  \\
5          & 10 & 0.15 & 0.2 & 5 & $\left[ {-3,1} \right]$ &                                   &                                   \\
\hline
\hline
$\surd $ &${\rm{ \times }}$           & $\surd $             & ${\rm{ \times }}$            & $\surd $             & $\surd $          & 62.26                             & \underline{84.56}                             \\
$\surd $ & ${\rm{ \times }}$           & ${\rm{ \times }}$             & $\surd $            & $\surd $             & $\surd $           & 61.92                             & 84.31                             \\
${\rm{ \times }}$ & $\surd $           & $\surd $             & ${\rm{ \times }}$            & $\surd $             & $\surd $           & \underline{62.93}                             & 84.49                             \\
${\rm{ \times }}$ & $\surd $           & ${\rm{ \times }}$            & $\surd $            & $\surd $             & $\surd $           & \textbf{63.57}                    & \textbf{84.62}                    \\
\hline
\end{tabular}
\end{table}

\begin{table}[!h]
\centering
\caption{The 3D detection performance of the different DES settings and above best GAS parameters on the KITTI validation split. The evaluation metric is mAP of 40 recall positions. Bolded and underlined values are the best and suboptimal performances for all methods, respectively.}
\label{table:table7}
\begin{tabular}{c|c|cc|cc|cc}
\hline
      $y_t$     &   $\tau_t$           & \multicolumn{2}{c|}{$d_t$ }         & \multicolumn{2}{c|}{$\tau_z$ } & \multirow{2}{*}{Ped. 3D} & \multirow{2}{*}{Car 3D}  \\
10         & 0.2 & 4 & 5 & $\left[ {-1.5,0.5} \right]$ & $\left[ {-3,1} \right]$ &                                   &                                   \\
\hline
\hline
$\surd $  & $\surd $             & $\surd $              & ${\rm{ \times }}$             & $\surd $          & ${\rm{ \times }}$          & 62.42             & 84.55                             \\
$\surd $  & $\surd $             & $\surd $              & ${\rm{ \times }}$             & ${\rm{ \times }}$          & $\surd $          & 63.08              & 83.21                             \\
$\surd $  & $\surd $             & ${\rm{ \times }}$             & $\surd $              & ${\rm{ \times }}$          & $\surd $          & \underline{63.57}              & \underline{84.62}                             \\
$\surd $  & $\surd $             & ${\rm{ \times }}$              & $\surd $              & $\surd $        & ${\rm{ \times }}$          & \textbf{63.64}      & \textbf{84.81}                    \\
\hline
\end{tabular}
\end{table}

\subsubsection{Experimental Analysis for SMS-PVRCNN}
SMS-PVRCNN uses the PV-RCNN backbone to perform experiments on the KITTI validation split, verifying the effectiveness of crucial parameters and modules. Since grid-based multi-view proposals are spatially aligned on BEV, the first-stage detector can directly calculate the multi-view consistency loss ${{\cal L}_{cons}}$ of CF-proposals without requiring the CKPS module. Specially, the following training experiments are conducted on the 15\% training split. As shown in Table \ref{table:table8}, using the same optimal settings as SMS-PointRCNN yields the highest 3D mAPs for pedestrians and cars. These results demonstrate that our multi-branch two-stage 3D object detection framework based on SMS and multi-view consistency constraints can exhibit consistent effectiveness on the point-based and grid-based backbones.

\begin{table}[!h]
\centering
\caption{The effectiveness of the crucial parameters and modules on the KITTI validation split for PV-RCNN backbone, with mAP of 40 recall positions. The above results are acquired by training on the 15\% training split. Bolded and underlined values are the best and suboptimal performances for all methods, respectively.}
\label{table:table8}
\setlength\tabcolsep{4pt}
\begin{tabular}{c|c|c|c|cc}
\hline
Pv1+Pv2+Pv3 & MVFP & ${{\cal L}_{cons}}$  & $\tau_z \in \left[ {-1.5,0.5} \right]$ & Ped. 3D & Car 3D  \\
\hline
\hline
${\rm{ \times }}$           & ${\rm{ \times }}$    & ${\rm{ \times }}$ & ${\rm{ \times }}$ & 49.45            & 81.94            \\
$\surd $             & $\surd $             & ${\rm{ \times }}$ & ${\rm{ \times }}$ & 51.51            & 82.30            \\
$\surd $             & $\surd $             & ${\rm{ \times }}$ & $\surd $          & \underline{51.88}            & \underline{82.33}            \\
$\surd $             & $\surd $             & $\surd $          & $\surd $          & \textbf{52.23}    & \textbf{82.93}   \\
\hline
\end{tabular}
\end{table}

\begin{figure}[!h]
	\centering
	\subfloat[]{
		\includegraphics[width=1.6in]{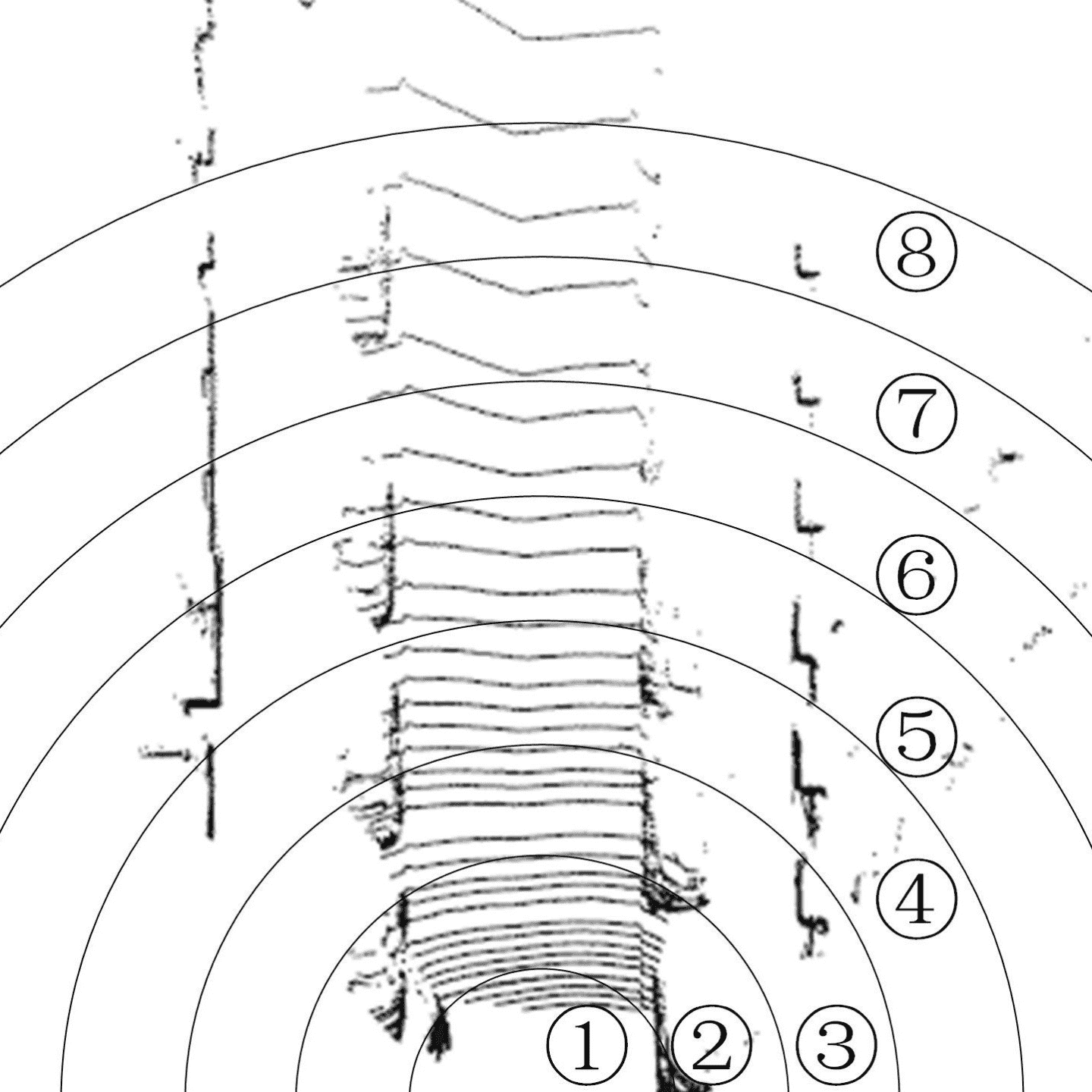}
		\label{fig:7a}}
	\hfil
	\subfloat[]{
		\includegraphics[width=1.6in]{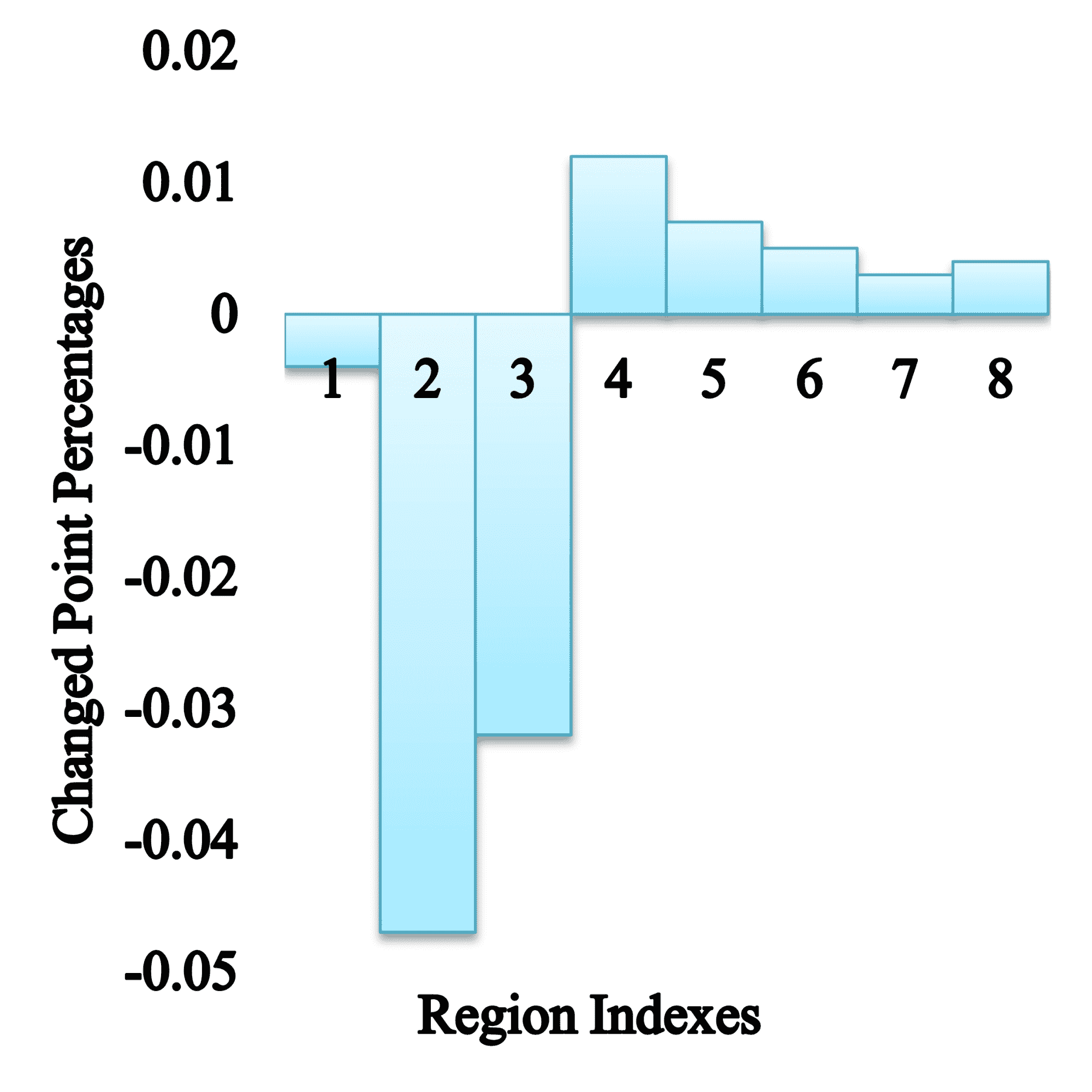}
		\label{fig:7b}}	
	\vspace{0.5cm} 
	\subfloat[]{
		\includegraphics[width=1.6in]{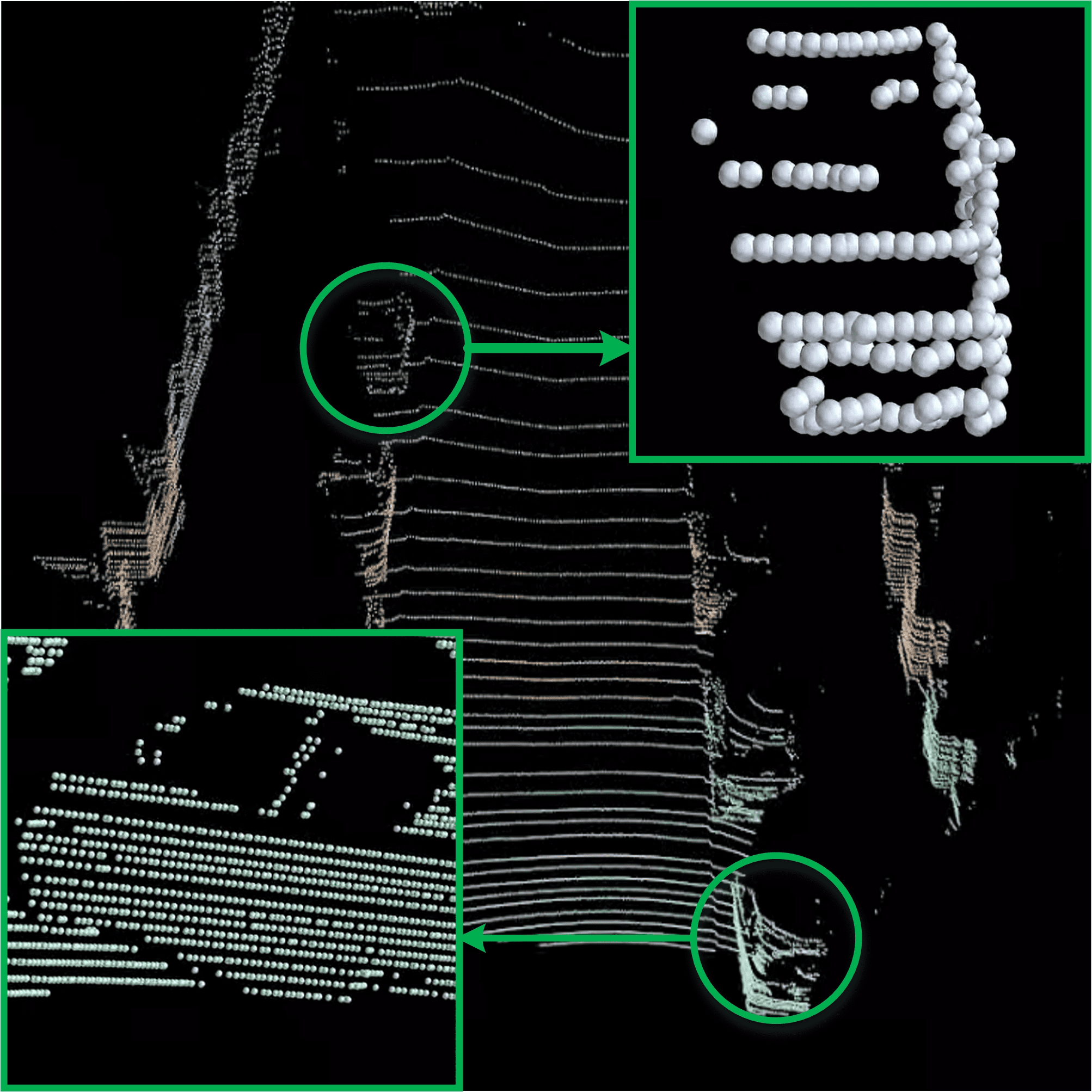}
		\label{fig:7c}}
	\hfil
	\subfloat[]{
		\includegraphics[width=1.6in]{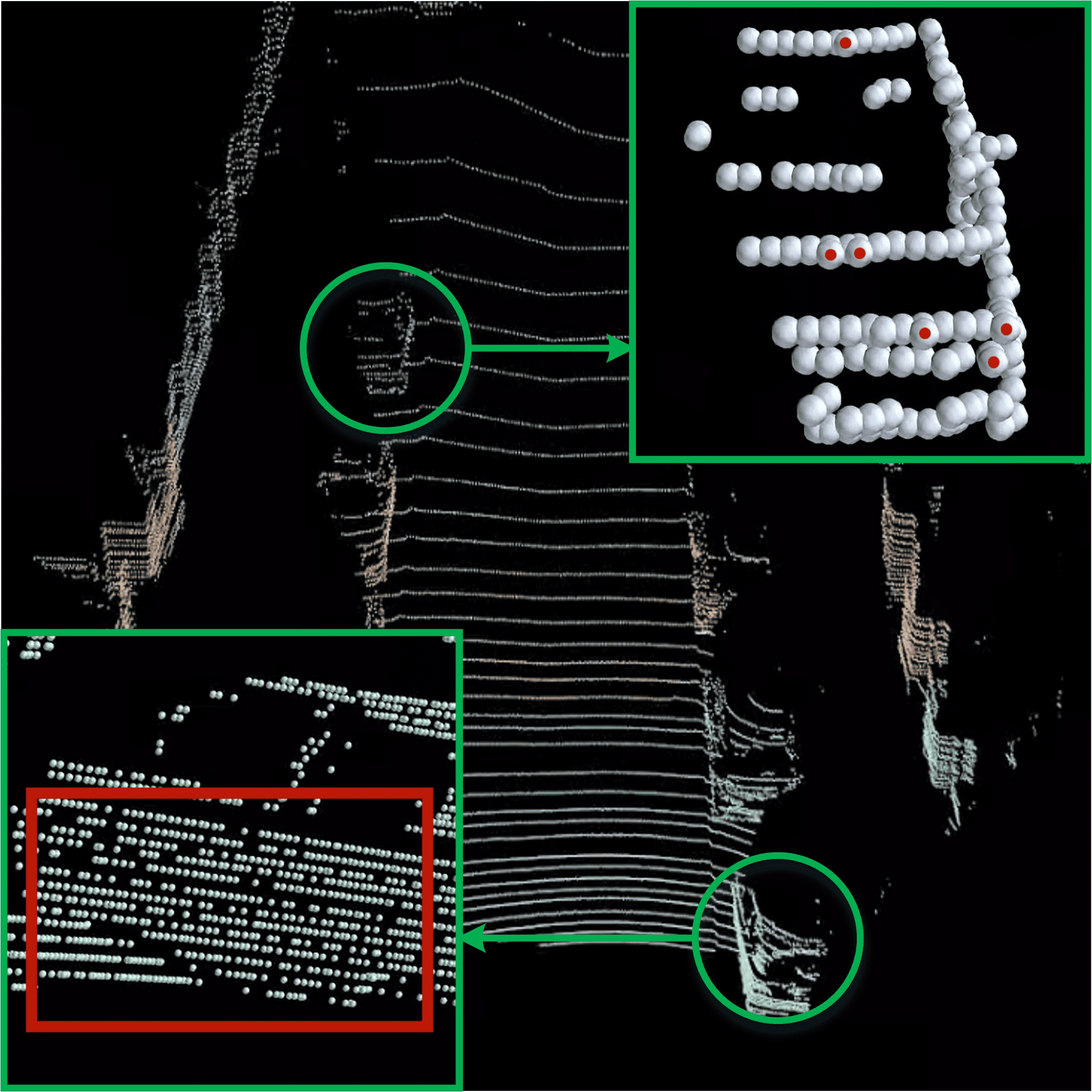}
		\label{fig:7d}}
	
	\caption{\revise{The statistical or visual graphs of DES on the KITTI dataset. (a) The regions of DES; (b) The changed point percentages in each region of DES; (c) The partial magnification graph of point cloud before DES; (d) The partial magnification graph of point cloud after DES with its point variations marked by red points or rectangles.
    }}
	\label{fig:figure7}
\end{figure}

\begin{figure*}[!htb]
\begin{center}
    \begin{tabular}{c@{\hspace{-4mm}}}
    \hspace{-15pt}\includegraphics[width=5.3in]{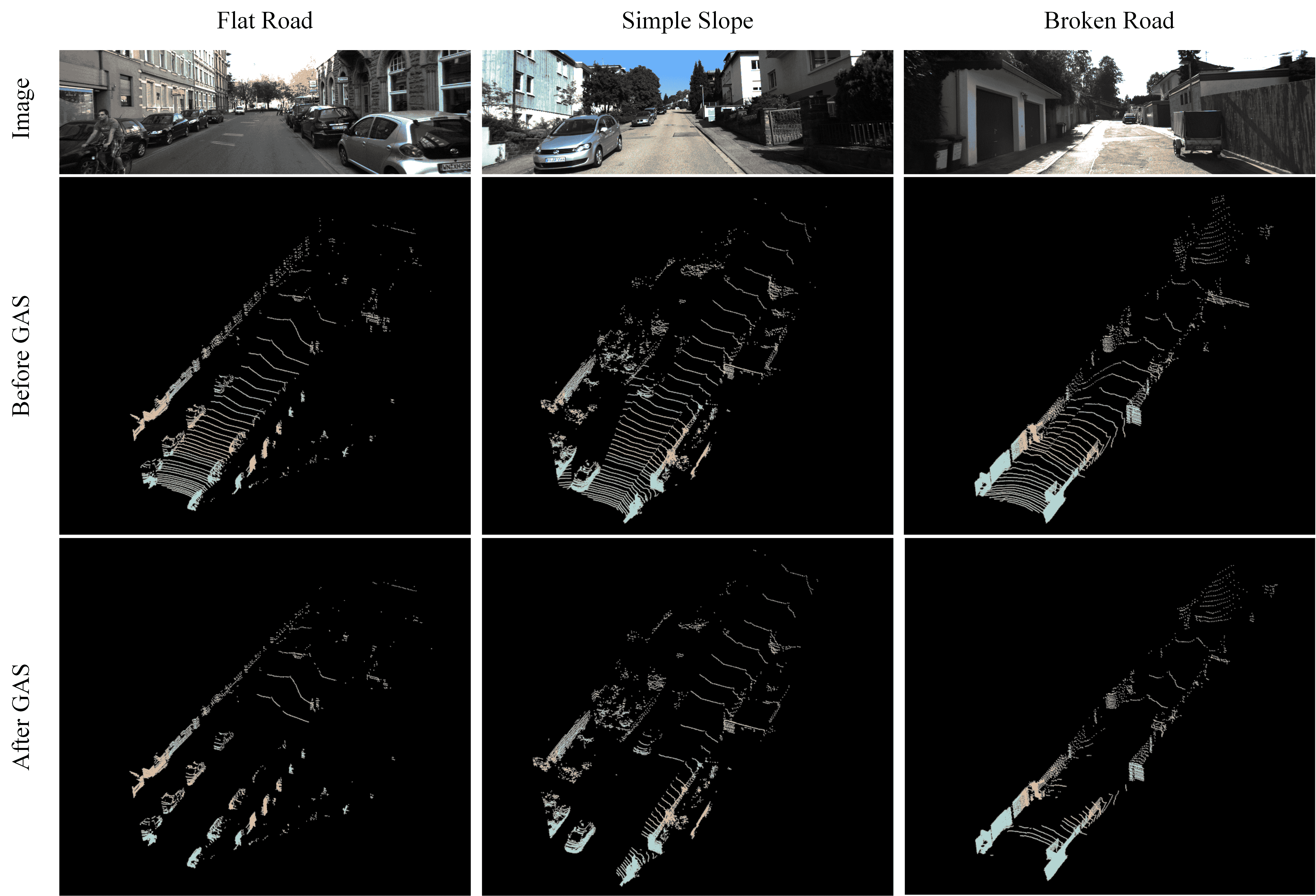}   \\
\end{tabular}
  \caption{\revise{Visual Graphs for the effective results of GAS on the KITTI dataset.}}
  \label{fig:figure_GAS1}
\vskip -10pt
\end{center}
\end{figure*}

\subsubsection{Subjective Analysis and Visualization of SMS module}
To further evaluate the effectiveness of our SMS module, we conduct statistical analyses and visual inspections on the KITTI dataset. DES divides the area into circular ring regions  \textcircled{1}$\sim$\textcircled{8} (see Fig. \hyperref[fig:figure7]{\ref{fig:figure7}(a)}), and adjusts the point percentages in different regions based on their densities. As illustrated in Fig. \hyperref[fig:figure7]{\ref{fig:figure7}(b)}, DES reduces the dense point percentages in near regions \textcircled{1}$\sim$\textcircled{3} while increasing the sparse point percentages in remote regions  \textcircled{4}$\sim$\textcircled{8}. The partial magnification graphs of point clouds before and after DES are shown in Fig. \hyperref[fig:figure7]{\ref{fig:figure7}(c)} and Fig. \hyperref[fig:figure7]{\ref{fig:figure7}(d)}, respectively, with variations in sampled points highlighted by red points or rectangle in Fig. \hyperref[fig:figure7]{\ref{fig:figure7}(d)}. Additionally, we assess the impact of GAS by visually comparing point clouds with and without ground points. As shown in Fig. \ref{fig:figure_GAS1}, the visual results indicate that GAS effectively removes a large number of ground points.

As illustrated in Fig. \hyperref[fig:figure7]{\ref{fig:figure7}(a)}, we define the circular ring regions \textcircled{1}$\sim$\textcircled{3} as the High-Density ($\mathcal{HD}$) region, while the circular ring regions  \textcircled{4}$\sim$\textcircled{8} are denoted as the Low-Density ($\mathcal{LD}$) region. As to the point counts within these two regions, we record the average variations of 50-frame point clouds after different sampling operations, as shown in Table \ref{table:table9}. In each region, the recorded variations include results from no sampling, random sampling, DES and GAS (namely $\mathcal{S}_0, \mathcal{S}_1, \mathcal{S}_2, \mathcal{S}_3$). In these two regions, the number ratios of  foreground points for $\mathcal{S}_i$ to $\mathcal{S}_0$ are defined as $R_1^{(\mathcal{HDS}_i)}$ and $R_1^{(\mathcal{LDS}_i)}$, and their calculation formulas are:
\begin{equation}
R_1^{(\mathcal{HDS}_i)} = \frac{N_{\mathcal{HDS}_i}^{(p_{fg})}}{N_{\mathcal{HD S}_0}^{(p_{fg})}},R_1^{(\mathcal{LDS}_i)} =  \frac{N_{\mathcal{LDS}_i}^{(p_{fg})}}{N_{\mathcal{LDS}_0}^{(p_{fg})}},i \in \left\{1,2,3 \right\}, \\
\end{equation}
where $N_{\mathcal{HDS}_i}^{(p_{fg})}$ and $N_{\mathcal{HDS}_0}^{(p_{fg})}$ are the foreground point counts for $\mathcal{S}_i$ and $\mathcal{S}_0$ in $\mathcal{HD}$. $N_{\mathcal{LDS}_i}^{(p_{fg})}$ and $N_{\mathcal{LDS}_0}^{(p_{fg})}$ are the same items in $\mathcal{LD}$. Next, we define the two region's number ratios of foreground points to all points for $\mathcal{S}_i$ as $R_2^{(\mathcal{HDS}_i)}$ and $R_2^{(\mathcal{LDS}_i)}$. These number ratios can be obtained by the following formulas:

\begin{equation}
R_2^{(\mathcal{HDS}_i)} = \frac{N_{\mathcal{HDS}_i}^{(p_{fg})}}{N_{\mathcal{HD S}_i}^{(p_{all})}},R_2^{(\mathcal{LDS}_i)} = \frac{N_{\mathcal{LDS}_i}^{(p_{fg})}}{N_{\mathcal{LDS}_i}^{(p_{all})}},i \in \left\{0,1,2,3 \right\}, \\
\end{equation}
where $N_{\mathcal{HD S}_i}^{(p_{all})}$ and $N_{\mathcal{LDS}_i}^{(p_{all})}$ are the numbers of all points for $\mathcal{S}_i$ in the two regions.

By analyzing Table \ref{table:table9}, we can see that both $\mathcal{S}_1$ and $\mathcal{S}_2$ lose some foreground points in the two regions. However, $\mathcal{S}_3$ increases the foreground point number, indicating that GAS can mitigate the loss of foreground points. Compared to random sampling, DES increases the foreground point percentages for enhancing distant objects in  $\mathcal{LD}$, with $R_1^{(\mathcal{LDS}_1)}$ increasing by 11.8\% over $R_1^{(\mathcal{LDS}_0)}$. Moreover, $R_2^{(\mathcal{HDS}_3)}$ is 14.6\% higher than $R_2^{(\mathcal{HDS}_0)}$, and $R_2^{(\mathcal{LDS}_3)}$ is 11.4\% higher than $R_2^{(\mathcal{LDS}_0)}$. The above improvements demonstrate the importance of GAS in focusing on foreground objects. All these results provide insights into the effectiveness of our SMS module.

\begin{table*}[!h]
	\centering
	\caption{The change of point number caused by different sampling operations in this paper on the KITTI dataset. The $R_1$ represents the number ratio of sampled foreground points to no sampled foreground points. The $R_2$ represents the number ratio of foreground points to all points for different sampling operations. }
    \label{table:table9}
	\begin{tabular}{c|c|c|c|c|c|c|c|c}
		\hline
		\emph{Point Num.} & $\cal {HDS}_\textbf{0}$ &$\cal {HDS}_\textbf{1}$ & $\cal {HDS}_\textbf{2}$  &$\cal {HDS}_\textbf{3}$ &$\cal {LDS}_\textbf{0}$ &$\cal {LDS}_\textbf{1}$ &$\cal {LDS}_\textbf{2}$ &  $\cal {LDS}_\textbf{3}$           \\
		\hline
        \hline
		$N_i^{p_{all}}$ & 14689  & 10419  & 9644            & 9972             & 4750   & 3426   & 4002            & 3789              \\
		$N_i^{p_{fg}}$   & 3681   & 2623   & 2416            & 3955             & 1409   & 1018   & 1184            & 1558              \\
		$R_1$          & -      & 71.3\% & 65.6\% & 107.4\% & -      & 72.3\% & 84.1\% & 110.6\%  \\
		$R_2$          & 25.1\% & 25.2\% & 25.1\%          & 39.7\%  & 29.7\% & 29.7\% & 29.6\%          & 41.1\%   \\
		\hline
	\end{tabular}
\end{table*}

\begin{figure*}[!h]
	\begin{center}
		\subfloat[PointRCNN]{
			\includegraphics[width=5.3in]{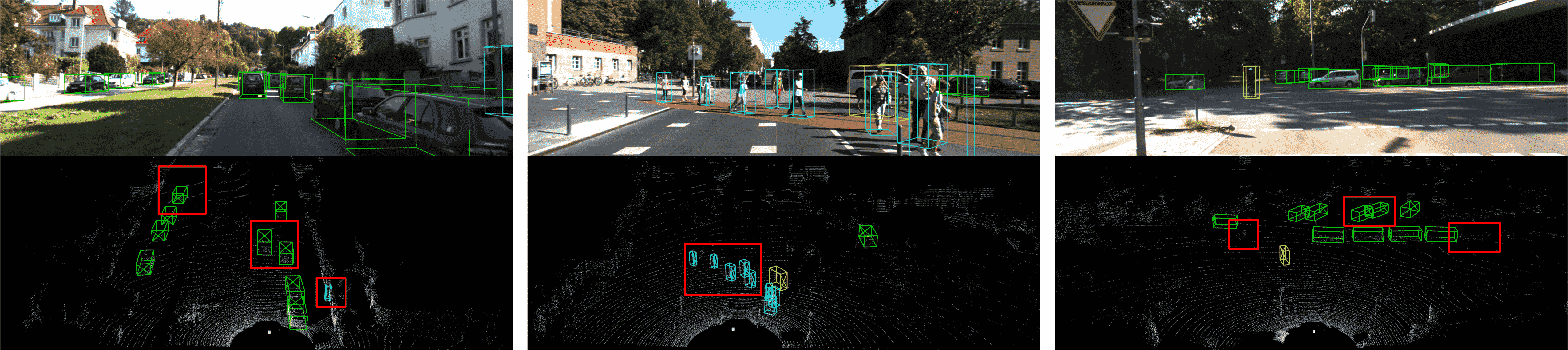}
			\label{fig:8a}}
		\vspace{-0.2cm} 
		\subfloat[SMS-PointRCNN]{
			\includegraphics[width=5.3in]{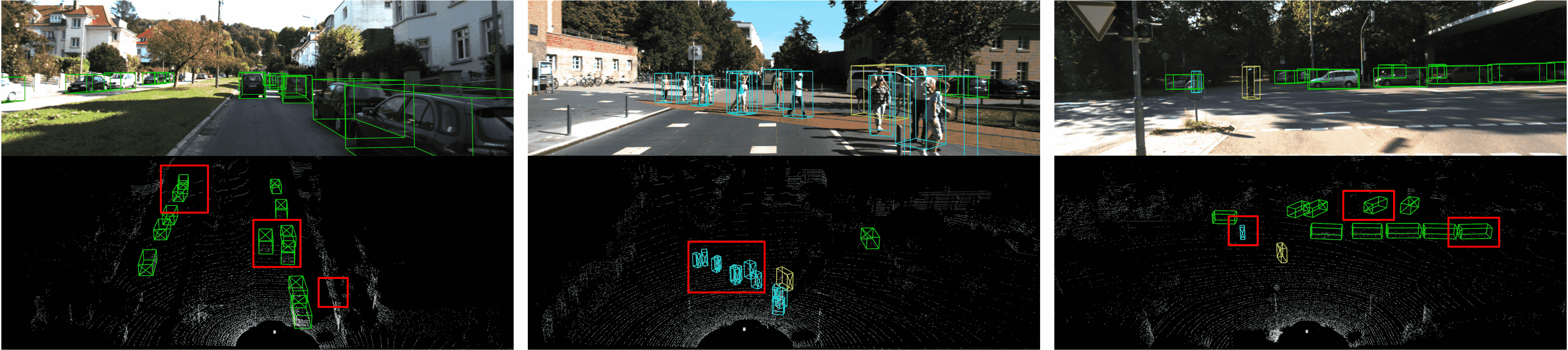}
			\label{fig:8b}}	
		\vspace{-0.2cm} 
		\subfloat[PV-RCNN]{
			\includegraphics[width=5.3in]{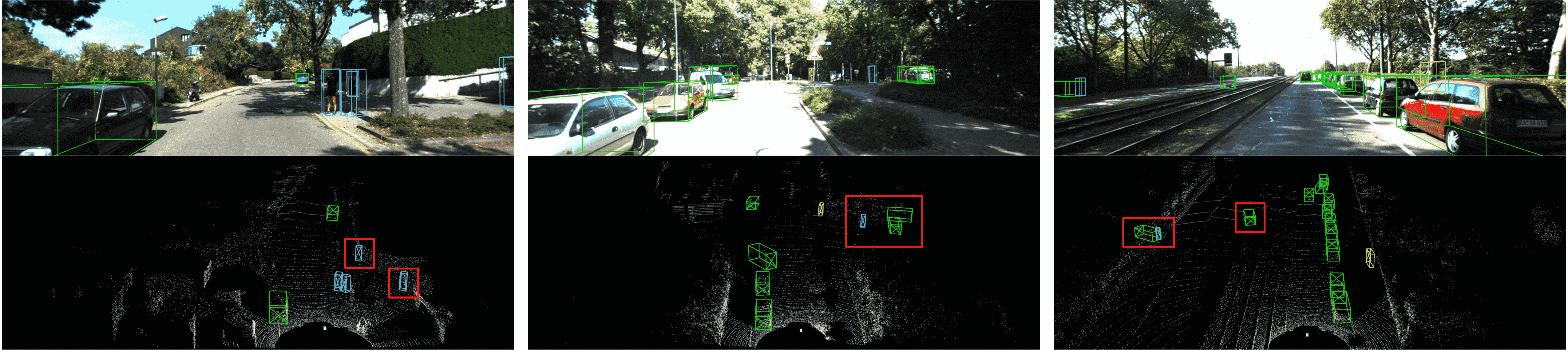}
			\label{fig:8c}}
		\vspace{-0.2cm} 
		\subfloat[SMS-PVRCNN]{
			\includegraphics[width=5.3in]{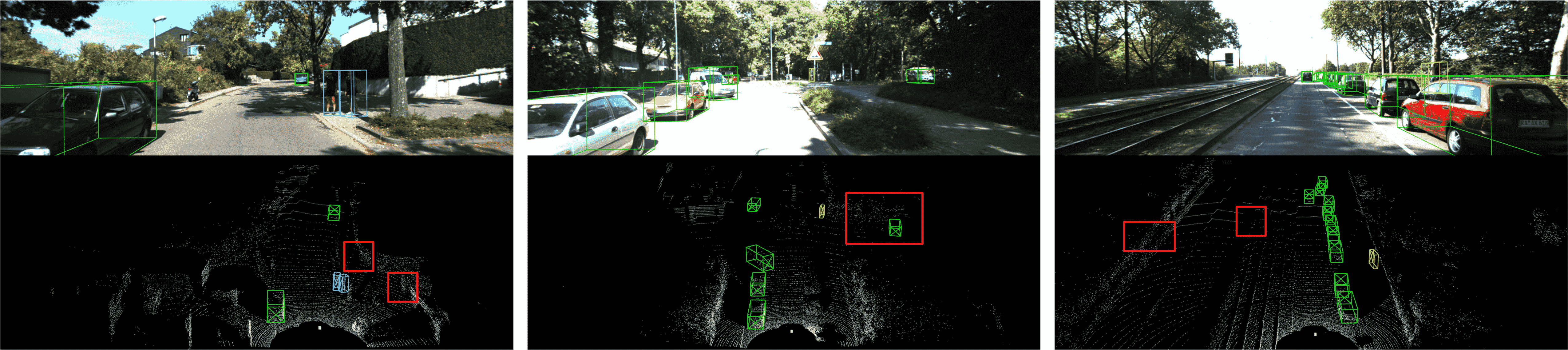}
			\label{fig:8d}}	
	
		\caption{Visual result comparisons between our methods and their backbones on the KITTI test split. In the same scene, the detection differences of each comparison are highlighted by red rectangles.}
		\label{fig:figure8}
	\end{center}
\end{figure*}

\subsubsection{Visual Comparisons of Detection Results}
The visual comparisons of detection results between our methods and their corresponding backbones are shown in Fig. \ref{fig:figure8} on the KITTI test split. Across these comparisons, our methods demonstrate fewer false positives than their respective backbones in the car and pedestrian classes. Notably, the visual result comparisons in Fig. \hyperref[fig:figure8]{\ref{fig:figure8}(a)} and Fig. \hyperref[fig:figure8]{\ref{fig:figure8}(b)} show that our SMS-PointRCNN achieves more accurate results for distant cars and small-sized pedestrians compared to PointRCNN. These findings illustrate our methods' ability to attain higher detection accuracy across a variety of backbones.

\begin{table}[!t]
\centering
\caption{\revise{Inference speed comparisons between different baselines and our method on the KITTI dataset, with Frames Per Second (FPS) as the primary evaluation metric.}}
\label{table:table10}
\begin{tabular}{c|c|c}
\hline
Method & Batch Size & FPS         \\
\hline
\hline
PointRCNN \cite{shi2019pointrcnn} & 2 & 9.5 \\
SMS-PointRCNN(ours) & 2 & 3.1 \\
\hline
PV-RCNN \cite{shi2020pv}* & 2 & 7 \\
SMS-PVRCNN(ours) & 2 & 2.4 \\
\hline
PV-RCNN++ \cite{shi2023pv}* & 2 & 9.25\\
SMS-PVRCNN++(ours) & 2 & 3.1 \\
\hline
\end{tabular}
\end{table}

\begin{figure*}[!htb]
\begin{center}
    \begin{tabular}{c@{\hspace{-4mm}}}
    \hspace{-15pt}\includegraphics[width=5.3in]{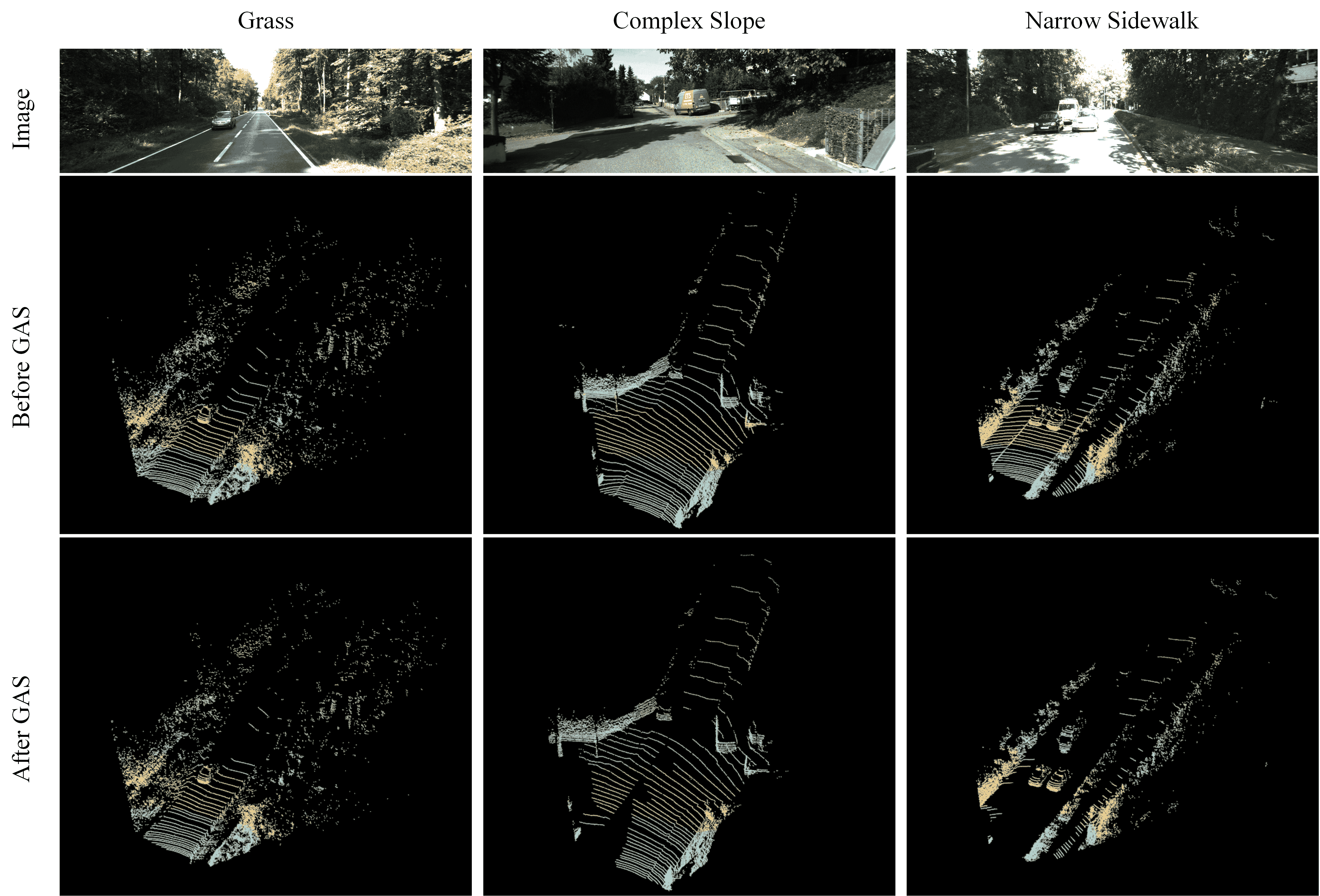}   \\
\end{tabular}
  \caption{\revise{Visual Graphs for the limited results of GAS on the KITTI dataset.}}
  \label{fig:figure_GAS2}
\vskip -10pt
\end{center}
\end{figure*}

\subsection{\revise{Discussion of Limitation and Future Work}}
\label{section:4D}
The proposed framework employs a multi-branch structure for 3D object detection. While this approach maintains model size by using shared network parameters, it substantially increases computational costs during multi-branch inference, resulting in slower inference speeds. \revise{As shown in Table \ref{table:table10}, our methods are approximately three times slower than several baselines.} To address this, future work will focus on more efficient utilization of multi-view point clouds to achieve both accuracy and faster inference. According to our current investigations, knowledge distillation presents a potential direction for enhancing efficiency in this regard.

\revise{Additionally, GAS adjusts the threshold for ground points by using the lowest point height in each rectangular grid, achieving effective results on flat roads, simple slopes, and broken roads (see Fig. \ref{fig:figure_GAS1}). However, as illustrated in Fig. \ref{fig:figure_GAS2}, GAS has limited results on more challenging terrains, such as grasses, complex slopes, and narrow sidewalks, where the primary ground plane or slope in each grid can exceed the set threshold, diminishing the effectiveness of ground abandonment. This limitation motivates future research to explore adaptive thresholding techniques that would enable GAS to better handle such challenging environments.}

\revise{While GAS may perform suboptimally under certain conditions, the overall robustness of our multi-branch framework helps to mitigate this limitation’s impact on performance. We believe that through further optimizations, including the potential use of a multi-view distillation framework, our approach can ultimately support real-time detection even with increased computational demands.}

\section{Conclusion}
This paper proposes a multi-branch two-stage 3D object detection framework based on a pluggable SMS module and multi-view consistency constraints. The SMS-based data preprocessing enhances point cloud representation by transforming the raw point clouds into multi-view points or voxels with distinct semantic emphases. The first-stage detector employs multi-branch parallel training to fully extract unique semantic features from these multi-view representations, while the proposal-level consistency constraints strengthen feature aggregation across multiple views. The second-stage detector obtains the multi-view fusion pooling features through the MVFP module, achieving more accurate 3D object detection. Experimental results demonstrate that our method significantly improves performance across various two-stage backbones, particularly enhancing weaker backbones like PointRCNN.

\bibliographystyle{IEEEtran}
\bibliography{bib}

\vspace{11pt}
\bf{}\vspace{-33pt}
\begin{IEEEbiography}[{\includegraphics[width=1in,height=1.25in,clip,keepaspectratio]{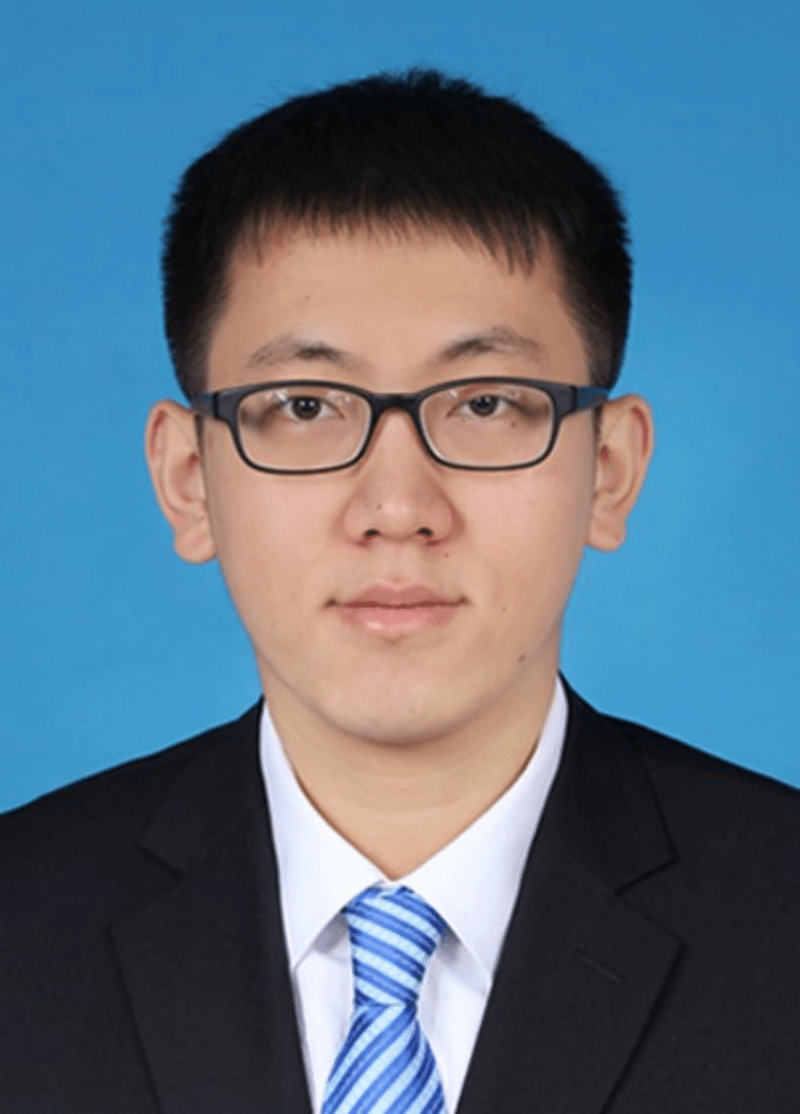}}]{Hao Jing}
received the B.E. and M.E. degree from University of Science and Technology Beijing, China, in 2017, where he is currently pursuing the Ph.D. degree in Taiyuan University of science and technology, China.  His research interests include 3D object detection and computer vision
\end{IEEEbiography}

\begin{IEEEbiography}[{\includegraphics[width=1in,height=1.25in,clip,keepaspectratio]{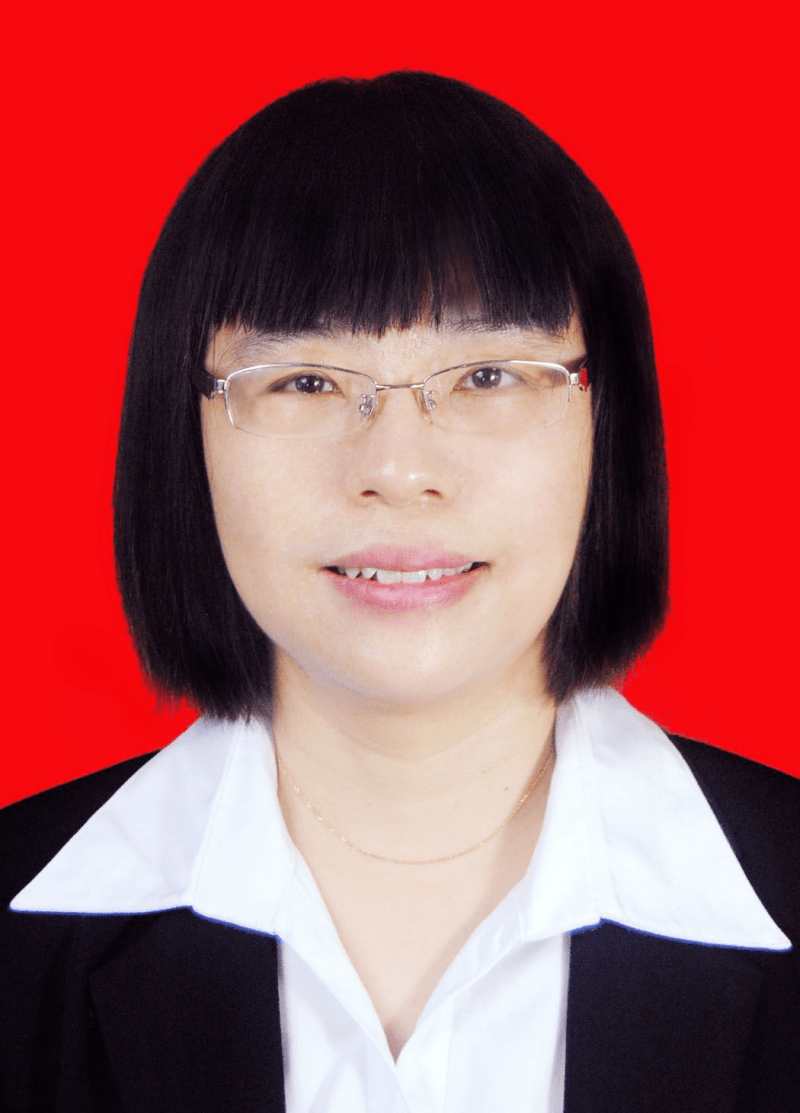}}]{Anhong Wang}
received B.S. and M.S. degrees from Taiyuan University of Science and Technology, China respectively in 1994 and 2002, and PhD degree in Institute of Information Science, Beijing Jiaotong University (BJTU) in 2009. She became an associate professor with TYUST in 2005 and became a professor in 2009. She is now the director of Institute of Digital Media and Communication, Taiyuan University of Science and Technology. Her research interests include image coding, video coding and secret image sharing, etc.
\end{IEEEbiography}

\begin{IEEEbiography}[{\includegraphics[width=1in,height=1.25in,clip,keepaspectratio]{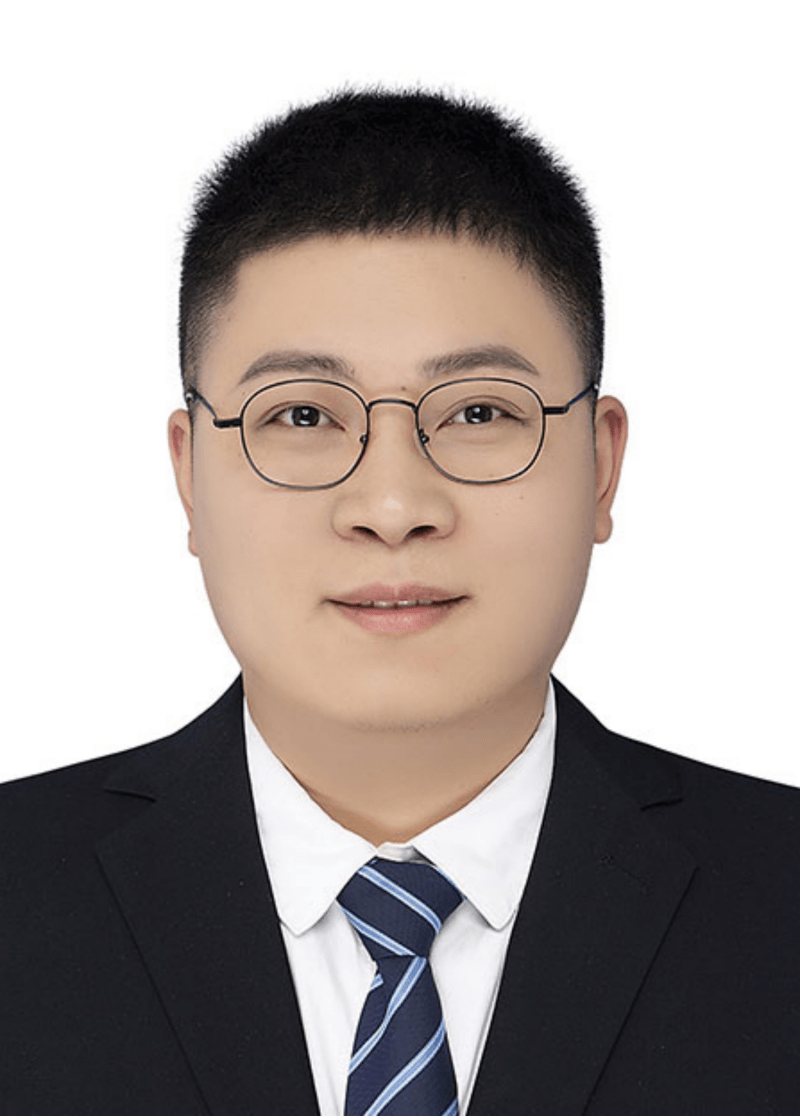}}]{Lijun Zhao}
received his M.S. degree from Taiyuan  University of Science and Technology in 2015 and PhD degree from Beijing Jiaotong University (BJTU) in 2019. He is currently an associate professor in the Institute of Digital Media and Communication, Taiyuan University of Science and Technology. His research interests include compressed sensing, image coding, multiple description coding, 3D video processing, image segmentation, etc.
\end{IEEEbiography}

\begin{IEEEbiography}[{\includegraphics[width=1in,height=1.25in,clip,keepaspectratio]{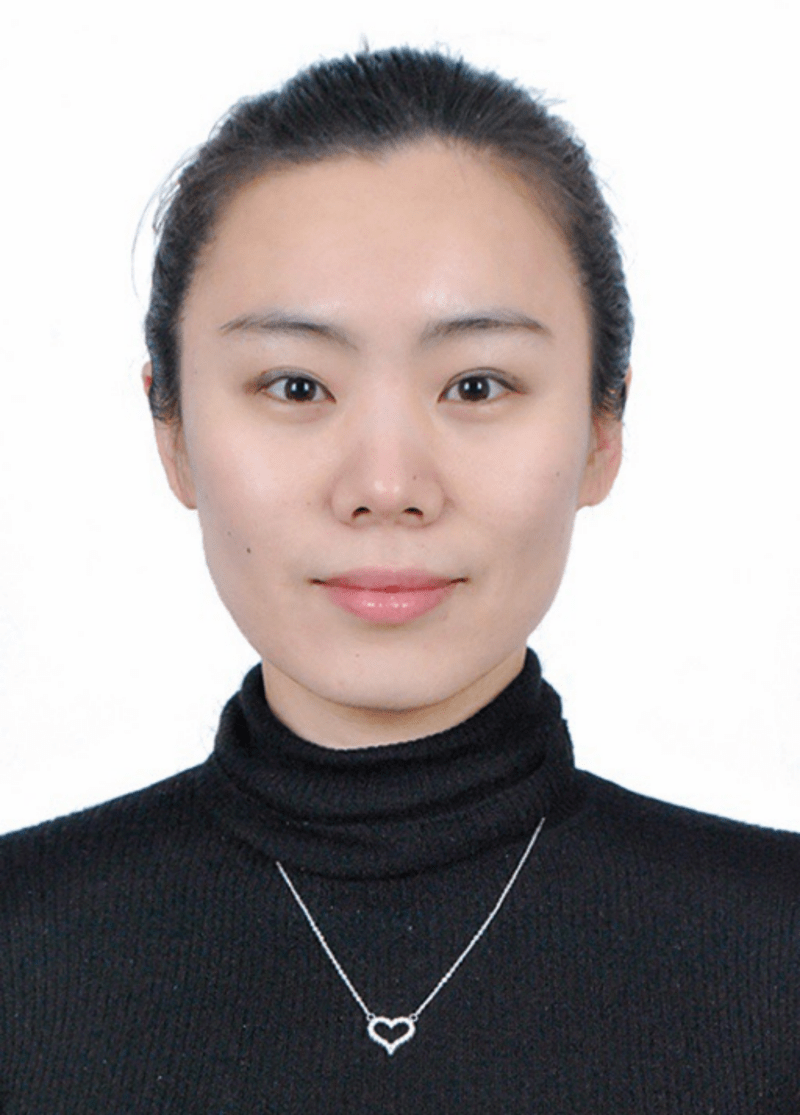}}]{Yakun Yang}
received the B.E. and M.E. degrees from Harbin University of Science and Technology and  Harbin Institute of Technology in 2012 and 2014,respectively,and the Ph.D. degree form Taiyuan University of Science and Technology in 2024. She is currently an lecturer with Taiyuan University of Science and Technology. Her research interests include 3D point cloud processing and computer vision.
\end{IEEEbiography}

\begin{IEEEbiography}[{\includegraphics[width=1in,height=1.25in,clip,keepaspectratio]{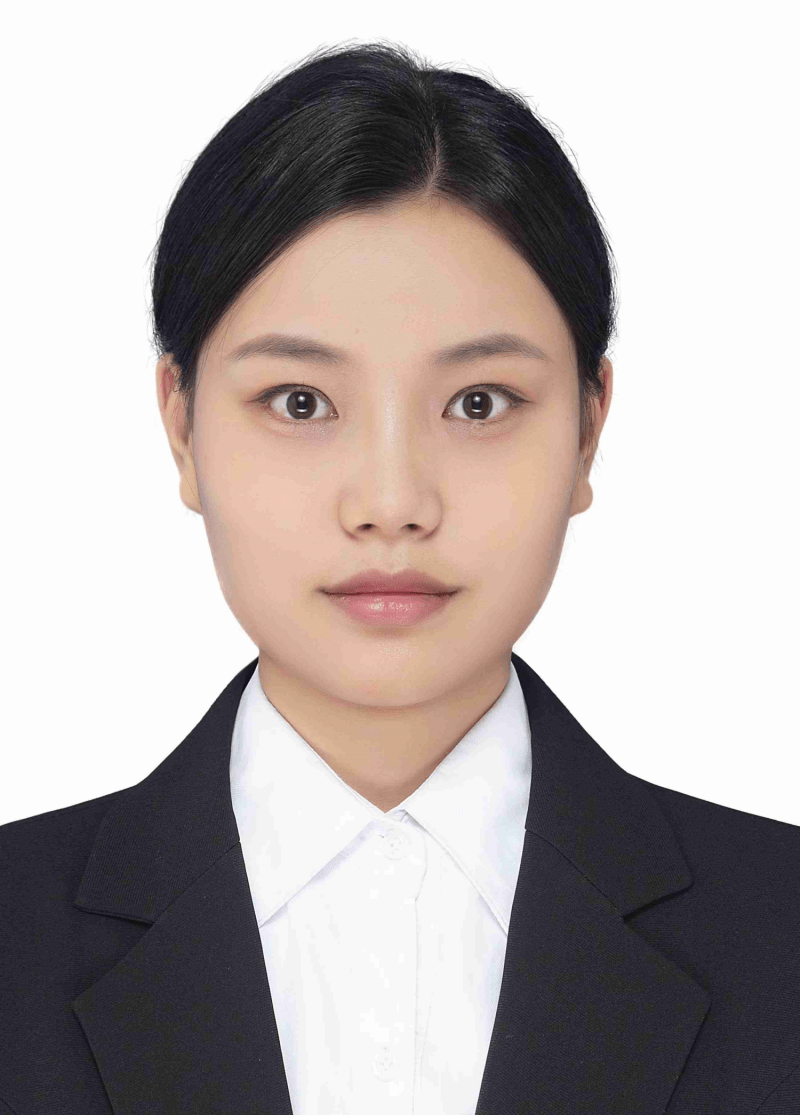}}]{Donghan Bu}
received the B.E. and M.E. degrees from Taiyuan University of Science and Technology, China, in 2017 and 2021, respectively. She is currently pursuing the Ph.D. degree in Taiyuan University of Science and Technology, China. Her research interests include 3D point cloud processing and computer vision.
\end{IEEEbiography}

\begin{IEEEbiography}[{\includegraphics[width=1in,height=1.25in,clip,keepaspectratio]{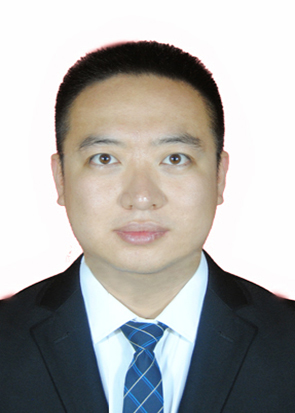}}]{Jing Zhang}
received the B.S. and M.E. degrees from North University of China and Taiyuan University of Science and Technology in 2014 and 2018, respectively, and the Ph.D. degree form Taiyuan University of Technology in 2023. He is currently an Lecturer with Taiyuan University of Science and Technology. His research interests include signal processing, emotion recognition, and video coding and transmission.
\end{IEEEbiography}

\begin{IEEEbiography}[{\includegraphics[width=1in,height=1.25in,clip,keepaspectratio]{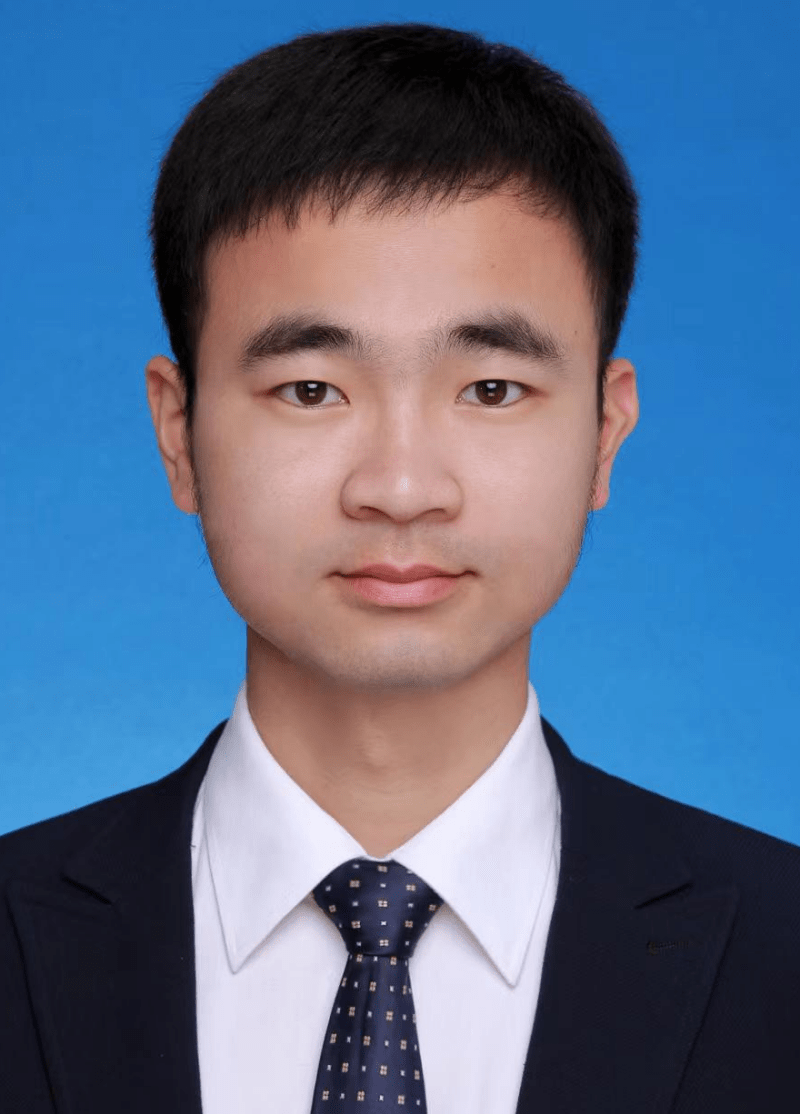}}]{Yifan Zhang}
	received the B.E. degree from the Huazhong University of Science and Technology (HUST), and the M.E. degree from Shanghai Jiao Tong University (SJTU). He is currently working toward the Ph.D. degree in the Department of Computer Science, City University of Hong Kong. His research interests include deep learning and 3D scene understanding.
\end{IEEEbiography}

\begin{IEEEbiography}[{\includegraphics[width=1in,height=1.25in,clip,keepaspectratio]{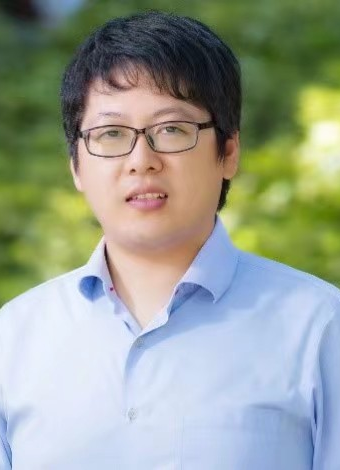}}]{Junhui Hou}
(Senior Member, IEEE)  is an Associate Professor with the Department of Computer Science, City University of Hong Kong. He holds a B.Eng. degree in information engineering (Talented Students Program) from the South China University of Technology, Guangzhou, China (2009), an M.Eng. degree in signal and information processing from Northwestern Polytechnical University, Xi’an, China (2012), and a Ph.D. degree from the School of Electrical and Electronic Engineering, Nanyang Technological University, Singapore (2016). His research interests are multi-dimensional visual computing.

Dr. Hou received the Early Career Award (3/381) from the Hong Kong Research Grants Council in 2018 and the NSFC Excellent Young Scientists Fund in 2024. He has served or is serving as an Associate Editor for \textit{IEEE Transactions on Visualization and Computer Graphics}, \textit{IEEE Transactions on Image Processing}, \textit{IEEE Transactions on Multimedia}, and \textit{IEEE Transactions on Circuits and Systems for Video Technology}. 
\end{IEEEbiography}

\vfill

\end{document}